\documentclass[journal]{IEEEtran}
\usepackage{color}
\usepackage{algorithm2e}
\usepackage{algorithmic}
\usepackage{pgfplots}
\usepackage{amsfonts}
\usepackage{amssymb}
\usepackage{mathtools}
\usepackage{bbm}
\usepackage{dsfont}
\usepackage{float}
\usepackage{hyperref}
\usepackage{comment}

\ifCLASSINFOpdf
\else
\fi
\begin{document}
%
\title{MRRC: Multiple Role Representation Crossover Interpretation for Image Captioning With R-CNN Feature Distribution Composition (FDC)} 
%
%
%

\author{Chiranjib~Sur \\ 
		Computer \& Information Science \& Engineering Department, University of Florida.\\
		Email: chiranjibsur@gmail.com
}

%
%

\markboth{Journal of XXXX,~Vol.~XX, No.~X, AXX~20XX}%
{Shell \MakeLowercase{\textit{et al.}}: Bare Demo of IEEEtran.cls for IEEE Journals}
%

\maketitle

\begin{abstract}
While image captioning through machines requires structured learning and basis for interpretation, improvement requires multiple context understanding and processing in a meaningful way.  
This research will provide a novel concept for context combination and will impact many applications to deal visual features as an equivalence of descriptions of objects, activities and events. 
There are three components of our architecture: Feature Distribution Composition (FDC) Layer Attention, Multiple Role Representation Crossover (MRRC) Attention Layer and the Language Decoder. FDC Layer Attention helps in generating the weighted attention from RCNN features, MRRC Attention Layer acts as intermediate representation processing and helps in generating the next word attention, while Language Decoder helps in estimation of the likelihood for the next probable word in the sentence. 
We demonstrated effectiveness of FDC, MRRC, regional object feature attention and reinforcement learning for effective learning to generate better captions from images. The performance of our model enhanced previous performances by 35.3\% and created a new standard and theory for representation generation based on logic, better interpretability and contexts. 
\end{abstract}


%
\IEEEpeerreviewmaketitle

\section{Introduction} \label{section:introduction}
\IEEEPARstart{B}{ig} data is struck with volume and variety of data and is creating huge challenges to merge them for analysis and inference. While large part of data is in images and videos and helping in better expression, entertainment and communication, it is difficult for machines to understand, merge, recommend and track the sentiments and trends of the specific and generalized happenings. To overcome this, captioning research must explore its fullest potential, for generalized and domain adaptation. 
Image captioning architectures \cite{sur2019survey} demonstrated diverse ways of effective sentence generation from visual features through various ways of representation generation and compositions from image features like Vgg \cite{Karpathy2015Deep}, ResNet (\cite{ren2015faster, Devlin2015Language, Gan2016}), Inception \cite{vinyals2015show} etc, mostly relying on object and attribute detectors to describe images (\cite{Karpathy2015Deep, Chen2015Mind, devlin2015exploring}) and later focused on attention based model (\cite{Xu2015Show, vinyals2015show, Mao2014deep, Devlin2015Language, yao2017boosting, rennie2017self, chen2018show}) and semantic factorization \cite{Gan2016}. 
Recent works with top-down objects from image regions (\cite{anderson2018bottom, lu2018neural}) has used hierarchical models. 
This architecture \cite{anderson2018bottom} has some serious drawbacks and we have introduced some solutions that are more robust and capable of introducing truly interpret-able representation instead of just relying on the network to learn them independently. Also, our architecture has much lower number of weights and can be trained in less than 20 epochs, while their architecture requires 50 epoch of rigorous training and then fine-tuning.  
Another drawback is related to global overview, derived out of average of all regional components. An overlap of a happy person and a sad one will produce only sad context or a negative way of representation because of the structural nature of the situation, which machine fails to interpret. Averaging suppresses many informative objects and overlapping average never works. The compositional characteristics of image-attention LSTM and intermediate transfer layer has limitations and creates a heuristic weighted selection (as $\textbf{a}$ in \cite{anderson2018bottom}) for image regions $\{v_1,\ldots, v_n\}$. Here, apart from a 2048 dimension LSTM, the intermediate transfer layer weights $\textbf{W}_b \in \mathbb{R}^{b \times d}$, $\textbf{W}_h \in \mathbb{R}^{b \times d}$, $\textbf{W}_a \in \mathbb{R}^{b \times d}$ are more than required. Hence, we curtailed the image-attention LSTM and $\textbf{W}_b$ and introduced $\textbf{W}_h \in \mathbb{R}^{b \times d}$ and $\textbf{W}_a \in \mathbb{R}^{b \times d}$ with much lower weights and this can help in propagation of information much faster. More details are provided in Section \ref{section:attention}. Our approach outperformed top-down approach in \cite{lu2018neural} which used only this image caption data for training, while our results were very close to top-down approach in \cite{anderson2018bottom}, which used three data (image captioning, VQA and Visual Genome Question Answering) to train their system for representation and use the features for different applications. Hence, direct comparison of our result with \cite{anderson2018bottom} will be unfair since we only use MSCOCO image caption dataset \cite{Gan2016}. We introduced a weighted image composition scheme to produce Feature Distribution Composition (FDC) through selection of the distribution of objects with the real global feature from the original image feature. This helps in better and quick selection of the weights for each corresponding regional object features. Also the intermediate transfer layer creates a fat structure, creating fan out situation for the information and further propagation of information to the image-attention LSTM layer get heavily diminished. Also, with lower gradient propagation, learning is slow and it is difficult to work with low value for initialization.  Also, in \cite{anderson2018bottom}, the intermediate self-attention feedback are the hidden layer outputs ($\textbf{h}_1$ and $\textbf{h}_2$) and most of the time propagates the initial generated errors, while the correct word embedding need to be propagate through the network for better learning. Overall, the model contains too many information fusion and it is difficult to control such mixing in a very heavy model with relatively lower resources (data and computation). We replaced these heavy weights with a simple and light weighted model and paid attention to composition of features from image features, global objects representation from the original image and use of semantic composition for construction of sentences. We showed that our composition of feature representation is a much advanced way of composition of representation that the system can distinguish and decode as sentences, while our proposed R-CNN feature composition architecture provides similar kind of performance, when experimented with MSCOCO dataset. 

The novelty of our work is multiple-fold. First, we advocate that crossover of representation can enhance image captioning, then we introduced a novel R-CNN features composition architecture for composition of structural representation in the form of Feature Distribution Composition (FDC) and lastly combined crossover representation, feature distribution composition and semantic composition for a better network. Lastly, we used SCST reinforcement learning scheme \cite{rennie2017self} for better performance enhancement and as a  moderating way of incorporating the cumulative topological effects of sequence of the sentence and generate changes in the network weights. 

The rest of the document is arranged with the details of theory of tensor product and its influence on our architecture in Section \ref{section:Tensor}, 
the description of the architecture and description of the different feature composition techniques in Section \ref{section:attention},
the intricacies of our experiments, results and analysis in Section \ref{section:results},
revisit of the existing works in literature in Section \ref{section-literature},
concluding remarks with future prospects in Section \ref{section:discussion}.

\section{Literature Review} \label{section-literature}
There are quite a number of works being done on image captioning like \cite{lu2018entity} CNN features and hash-tags from users as input, \cite{lu2018neural} with sentence `template', \cite{you2018image} sentiment-conveying image descriptions. \cite{melnyk2018improved} reported the comparison of context-aware LSTM captioner and co-attentive discriminator for image captioning. \cite{wu2018joint} used question features and image features, \cite{chen2017structcap} parsing tree StructCap, \cite{jiang2018learning} sequence-to-sequence framework, and  \cite{wu2018modeling} dual temporal modal, 
Image-Text Surgery in \cite{fu2018image}, \cite{chen2018show}  attribute-driven attention, \cite{cornia2018paying} generative recurrent neural network, \cite{zhao2018multi} MLAIC for better representation. Also there is \cite{li2018coco} text-guided attention, \cite{chen2017reference} reference based LSTM, \cite{chen2017show} adversarial neural network,  high-dimensional attentions \cite{ye2018attentive},  
\cite{wang2017skeleton} coarse-to-fine skeleton sentence, \cite{chen2018factual} specific styles,  
\cite{chen2018groupcap} structural relevance and structural diversity,  multimodal attention \cite{liu2017mat}, \cite{harzig2018multimodal} popular brands caption
\cite{liu2018show} diversified captions, 
\cite{chunseong2017attend} stylish caption, 
\cite{sharma2018conceptual} sub-categorical styles, 
\cite{yao2017incorporating} personalized captions,  
\cite{zhang2017actor} studied actor-critic reinforcement learning,  
\cite{fu2017aligning} scene-specific attention contexts, 
\cite{ren2017deep} policy network for captions,  
\cite{liu2017improved} reinforcement learning based training,   
\cite{cohn2018pragmatically}  distinguish between similar kind for diversity,  \cite{liu2017attention}  improved with correctness of attention in image,   
\cite{lu2017knowing} adaptivity for attention,  
\cite{vinyals2017show} used combination of computer vision and machine translation, 
\cite{zhang2018more} used adaptive re-weight loss function,   
\cite{park2018towards} personalized captioning,   
\cite{wu2017image} high level semantic concept,   
\cite{vinyals2015show} used visual features and machine translation attention combinations,     
\cite{Gan2017Stylenet} different caption styles,  
\cite{Jin2015Aligning} shifting attention,  
\cite{Kiros2014Unifying} characteristics of text based representations, 
\cite{Pu2016Variational} variational autoencoder representation, 
\cite{Socher2014Grounded} dependency trees embedding, 
\cite{Sutskever2011Generating}  character-level language modeling,  
\cite{Sutskever2014Sequence}  fixed dimension representation,  
\cite{LTran2015Learning}   3-dimensional convolutional networks, 
\cite{Tran2016Rich} human judgments, out-of-domain data handling,   
\cite{You2016Image}  semantic attention,   
\cite{Gan2016}  Semantic Compositional Network (SCN), 
\cite{Girshick2014}  localize and segment objects,  
\cite{Jia2015}  extra semantic attention,  
\cite{Kulkarni2013} content planning and recognition algorithms,  
\cite{Kuznetsova2014}  new  tree  based  approach  to composing expressive image descriptions, 
\cite{Mao2015} transposed weight sharing scheme,  
\cite{Mathews2016}  different emotions and sentiments,   
and \cite{Yang2011} where  nouns,  verbs,  scenes  and  prepositions  used for structuring sentence.

\section{Image Weighted Tensor Crossover \& Representations Theory} \label{section:Tensor}
The concept is motivated by multiple extraction of different aspects of the images and combining them, in absence of enough trigger neurons for many unattended part of the image features. Uniformly distribution of triggering of every sector of the image is also not feasible. Hence, we pretend that this overall combinations is an approximation event for neural composition network function. Mathematically, we can say for image $\textbf{I}$, we define an approximate context function $\Phi(.)$ through the combination of individual neuron triggered transformation of the image regions defined as $\sigma(\textbf{W}_1\textbf{I} + \textbf{b}_1)$, $\sigma(\textbf{W}_2\textbf{I} + \textbf{b}_2)$ and so on. The structural component of features in images are flattened through \cite{ren2015faster}.

\subsection{Approximation}
Approximation has always lead to ease of computation, but here we are focused on approximation for scalability. For better approximation and compositional characteristics of representations, our architecture learns to predict two near-optimum representation tensors that can be combined to generate the maximum likelihood representation. The purpose is to complement each other, but in case of presence in both representations,  non-linearity can neutralize such effects. These near-optimum tensors are learned from data and are generated as a function of visual features $\textbf{v}$ for image $\textbf{I}$. For example, we consider the time-independent representation $\textbf{R}$ in Equation \ref{eq:TPR}, 
\begin{equation} \label{eq:TPR}
 \textbf{R} =  f_1(\textbf{v})f_2(\textbf{v})
\end{equation}
with functions $f_1(.)$ and $f_2(.)$, we can also include time and rewrite representation $\textbf{R}_t$ as, 
\begin{equation} 
 \textbf{R}_t = \sum\limits_{i=0}^{i=t} c_i p_i^T
\end{equation}
where we interpret functions $f_1(.)$ and $f_2(.)$ as $c_i$ and $p_i$ and are the context and positional significance vector for the image and is used to generate partial global representation $c_i p_i^T$ for each word in the sentence (or elements in a topological sequence), where the word embedding vector is defined as $c_i$ and the positional or semantic interpretation is provided as $p_i$. 
We retrieve back $c_i$ from decoder, since we have defined likelihood and we know $\textbf{R}_t$ as the structure of sentence with positional significance $\{p_1,p_2,\ldots,\}$ for each word $\{c_1,c_2,\ldots,\}$. Imposed orthogonality generates the detection theory for decoding as $p_i^Tp_j=0 \mid \forall \textbf{ } i \neq j$ and $p_i^Tp_j=1 \mid \forall\textbf{ } i = j$, if we know $c_i$. The approximate structure of the function creates retrieval as $c_1 p_1^Tp_1$, $c_2 p_2^Tp_2$, $c_3 p_3^Tp_3$ and so on from each $\textbf{R}_t$. Equivalently, we can write, estimated $c_i$ as $\widehat{c}_i$,
\begin{equation}
 \widehat{c}_i = \textbf{R}_t p_i = c_ip_i^Tp_i
\end{equation}
where $\textbf{R}_t$ is Multiple Role Representation Crossover (MRRC). To apply this concept for language generation, we have two situations. One is where we know and defined one set of vectors and we can retrieve the other,  with the help of $\textbf{R}_t$ from contexts (here images). Another is when we derive both the vectors and $\textbf{R}_t$ from from contexts (here images) and retrieve the other vectors. Experiments show that the latter is slightly better and this is due to approximations. 

For image caption generation problem, it is difficult to retrieve whole $\textbf{R}_t$ from $\textbf{I}$ and hence we can define an approximate scheme where we compose partial $\textbf{R}_t$ as $f(\textbf{v})$ and add it to $\textbf{R}_{t-1}$ for partial sequence context and establishing the topological dependency and thus retrieve context word $c_i$ simultaneously from the image as attention. This architecture can be represented mathematically as the following conditional probability equation. 
\begin{equation} 
\begin{aligned}
 \mathcal{P}(c_i \mid f(\textbf{v}), c_1,c_2,\ldots) & =   \mathcal{P}(c_i \mid f(\textbf{v})) + \mathcal{P}(c_i \mid c_1,c_2,\ldots) \\
 & = \mathcal{P}(c_i \mid f(\textbf{v})) + \mathcal{P}(c_i \mid \textbf{R}_{t-1})
\end{aligned}
\end{equation}
We have $\mathcal{P}(c_i \mid f(\textbf{v}), c_1,c_2,\ldots)$ as the likelihood representation, attended as $\textbf{R}_t$. 
The overall mathematical interpretation  can be written as the following equation. 
\begin{equation}
\begin{aligned}
 \textbf{R}_t &= c_1 p_1^T + c_2 p_2^T + \ldots + f(\textbf{v}) \\
 &= \sum\limits_{i=0}^{i=t} c_i p_i^T  + f(\textbf{v})
\end{aligned}
\end{equation}
where $f(\textbf{v})$ can be regarded as structural correction by image attention. The ideal situation would have been when $c_t p_t^T = f(\textbf{v})$ or $c_t p_t^T = f(\textbf{v} \mid \textbf{h}_{t-1})$ and is denoted as the following equation. 
\begin{equation} 
\begin{aligned}
\textbf{R}_t  &= c_1 p_1^T + c_1 p_1^T + \ldots + c_t p_t^T \\
       &= \sum \limits_{i=1}^{i=t} c_i p_i^T 
\end{aligned}
\end{equation}
where we know the sentence $\{w_1,w_2,\ldots,w_n\}$ (for image $\textbf{I}$) and we can easily retrieve $\textbf{R}_t$ by adding $c_i p_i^T$ to $\textbf{R}_{t-1}$ from image and retrieving back $c_i$ as $c_i p_i^Tp_i$ (as $p_i^Tp_i = 1$) through a sequence decoding technique. So apparently, we are getting the next word $c_i$ from image $\textbf{I}$ but through $\textbf{R}_{t-1}$, but after each step we only know $w_i$ at time $t$ and before time $t$ based on the decoded sequence, instead of all $w_i$ before and after time $t$. 
This strategy will solve many problem related to generation of a complex and combined representation for the sentence and helps in scaling up the context diversity and generation capability. Since this is a non-linear transformation that is made to behave like reverse estimation, we will also predict the positional or semantic interpretation representation, an equivalent to topological information for a sequence. 
If time factor is considered, the initial projection of the tensors goes through the following series of updation, 
\begin{itemize}
 \item $\textbf{R}_0 \leftarrow Null $
 \item $\textbf{R}_1 \leftarrow \textbf{R}_{0} + f_1(\textbf{v})$
 \item $New\_Word \leftarrow w_1 = \textbf{R}_1 * p_1$
 \item $\textbf{R}_{2} \leftarrow \textbf{R}_{1} + w_1*p_1 + f_1(\textbf{v})$
 \item $\textbf{R}_{t+1} \leftarrow \textbf{R}_{t} + w_t*p_t + f_1(\textbf{v})$
\end{itemize}
where $p_i$ is defined or trained from the data and $\{w_1,w_2,\ldots,w_n\}$ is sentence for image $\textbf{I}$. This idea is converted into a form of structural neural function, where, we train neural network to define the functions.  

\subsection{Divide and Conquer Strategy In Deep Learning}
The fundamental difference between "divide and conquer" strategy and neural ensemble strategy is that the former is focused on combination generation instead of utilization individually as concatenated series. Here, our approach is to divide the situation and later combine the inference, instead of expecting details in single function. 
While approximation is bounded through mathematical structures, the details of the functional approximation is operated in a "divide and conquer" algorithm, where, instead of relying on some weights to generalize well for all the images, we diversify the work to different weights for better generalization. This is a perfect example of "divide and conquer" strategy in deep neural network, mainly concentrating on task of decoding and related to sequence generation and processing. Sequence processing is computation intensive and contextual segregation is inevitable for proper generation of topologically significant sequences. While, most approaches are dependent on weighted transformation, the models lack generalization for sequence generalization and hence, we introduce the idea of Multiple Role Representation Crossover for better learning of usefulness  from images. Mathematically, we can define this "divide and conquer" strategy in deep neural network as the following.
\begin{equation}
 \textbf{y} = \Phi(\textbf{x}) = \Phi(\Psi_1(\textbf{x}),\Psi_2(\textbf{x}), \ldots, \Psi_n(\textbf{x}))
\end{equation}
where we have defined $\Phi(.)$ as the transformation function of pre-likelihood and $\Psi_i(.)$ as the $i^{}$th intermediate extraction of contexts. Instead of characterizing each function $\Psi_i(.)$ for specific purposes, we define $\Psi_i(.)$ as component function that take their required role and is trained to that for different situations. 

\subsection{Local Recurrent Strategy In Deep Learning}
Local Recurrent Strategy is defined as identification of the most relevant and useful context (local context) in the search space instead of processing the whole context (global context). In this work, we have devised this approach for our strategy and is more like the swarm intelligence strategy, where the swarm always tries to follow some local best for improvement instead of global ones. Several local contexts will lead to the sentence, instead of learning a global context for the sentence. In this work, for sentence generation, the purpose is identification of some of the best local structural context for proper generation of the words for the sentence and these are derived from lower level image information. Mathematically, we can consider the following equation as one kind of Local Recurrent Strategy. 
\begin{equation}
 \textbf{y}_t = \Phi(\textbf{I}) = \Phi(\textbf{A}_t\textbf{I})
\end{equation}
where we have defined $\Phi(.)$ as the transformation function to pre-likelihood, the combination of local contexts $\textbf{A}_t\textbf{I}$ as context for the sentence at time $t$ through the adjustment matrix $\textbf{A}_t$. 
This can be regarded as Local Recurrent Strategy for the recurrent decoding, where the topology is established through variables at the language decoder level. Local Recurrent Strategy helps in diversification of contexts through "context shifts" strategies and is generated from the image features directly. Lastly, Local Recurrent Strategy helps in estimation of contexts that are not biased that is not learned through bias in the language decoder LSTM, where the learning of the language decoder plays important role.

\section{Attention Composition Architectures} \label{section:attention}
\begin{figure*}
\centering 
\includegraphics[width=\textwidth]{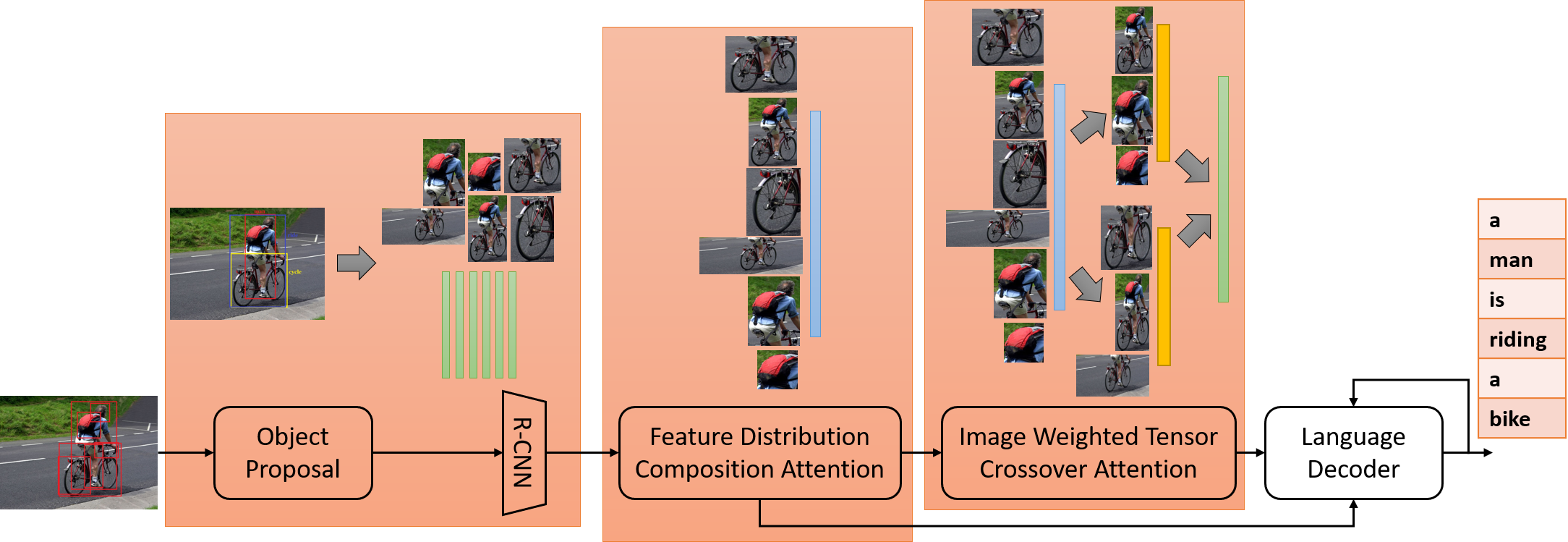}  
\caption{Overall FDC-MRRC Architecture With FDC, IWTC and RCNN Feature Fusion. }
\label{fig:R5}
\end{figure*}

The main motivation of introducing this architecture is the lack of previous architectures in defining a model that can capture different aspects of the situation. Attention is not generalized. But with series of local contexts can approach generalization and is a much powerful tool.  
Our architecture is composed of three components: Feature Distribution Composition (FDC) Attention, Multiple Role Representation Crossover (MRRC) Attention and Semantic Composition based Language decoder. FDC helps in feature composition from images, MRRC helps in selection through prediction of the next words through generation of semantics for the decoder LSTM layer. Different concepts of attentions are introduced and fused in the architecture that can generate the perfect combination for image captions. 
Figure \ref{fig:R5} provided a diagrammatic overview of the architecture and subsequently, we provided more details of the individual components.

\subsection{Feature Distribution Composition Attention}
\begin{figure}
\centering 
\includegraphics[width=.5\textwidth]{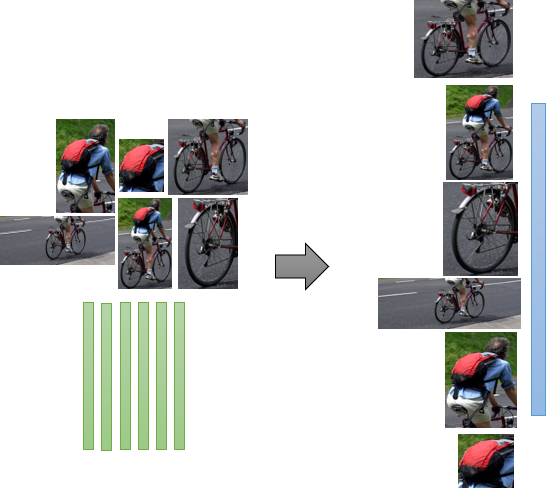}  
\caption{Feature Distribution Composition (FDC) Attention Layer.}
\label{fig:FDC_generator}
\end{figure}

Feature Distribution Composition Attention is characterized for detention of the semantic composition from the regional objects features and the semantic composition expresses more with more regularization of the lower level extracted features. Previous works on this kind of architecture utilized lower level CNN based feature of images for caption generations. CNN feature provides a limited and selective overview of the images and the composition cannot be varied for the network to be leveraged upon. But, RCNN features consist of series of features in individual forms that can heuristically be selected to compose a constructive component. Individually, the RCNN features are much more sparse, but the weighted component composition can be much more meaningful and constructive. We define an effective and efficient way of weighted selection scheme and call it Feature Distribution Composition (FDC) Attention Layer. Compared to the bottom up approach \cite{anderson2018bottom}, it is much more effective and lower in the number of weights and this scheme helped in out-perform previous works like \cite{lu2018neural}. This architecture will perform much better than \cite{anderson2018bottom}, if compared on the same platform, as \cite{anderson2018bottom} is trained with data from multiple sources. Mathematically, Feature Distribution Composition (FDC) Attention, denoted as $\widehat{v}_t$, can be derived by the following set of equations. 
\begin{equation}
 \overline{v} = \frac{1}{k} \sum\limits_{i=1}^{i=k} v_i
\end{equation}
where $\{v_1,v_2,\ldots,v_k\}$ are the regional CNN features and $v_i \in \mathbb{R}^{2048}$ $\forall$ $i \in \{1,2,\ldots,k\}$. The initial parameters of the LSTM for the language decoder is important as the hidden states are related to establishment of the topological relationship and time and we initialized them as the followings, considering that the FDC is related to the top overview of the image. 
\begin{equation}
 \textbf{h}_{0}, \textbf{ } \textbf{c}_{0} = \textbf{W}_{h_0}\overline{v}, \textbf{W}_{c_0}\overline{v}
\end{equation}
where we have defined $\textbf{a}_{t} \in \mathbb{R}^{b \times d}$, $\textbf{a}_{t} \in \mathbb{R}^{b \times d}$ for transformation of the features.
The next task of this model is development of the Intermediate Transfer Layer 
\begin{equation} \label{eq:st0}
 \textbf{a}_{t} = \textbf{W}_{a} \tanh (\textbf{W}_{h} \textbf{h}_{t-1})
\end{equation}
where $\textbf{a}_{t} \in \mathbb{R}^{k \times 1}$, 
$\textbf{W}_{a} \in \mathbb{R}^{m_x \times k}$, 
$\textbf{W}_{h} \in \mathbb{R}^{d \times m_x}$, 
$k$ is the number of considered maximum regional features in images. 
\begin{equation}
 \alpha_t = \mathrm{softmax}(\textbf{a}_t)
\end{equation}
where $\sum\limits_{i=1}^{i=k} \alpha_{i,t} = 1 $ and $\alpha_t \in \mathbb{R}^{k \times 1}$ and the Feature Distribution Composition Attention $\widehat{v}_t$ is provided as a summation of different regions to compose the attention. 
\begin{equation} \label{eq:en0}
 \widehat{v}_t =  \sum\limits_{i=1}^{i=k} v_i \alpha_{i,t}
\end{equation}
where we have $\widehat{v}_t \in \mathbb{R}^{b \times 2048}$ where $b$ is the batch size and $d$ is the hidden layer dimension. 
Figure \ref{fig:FDC_generator} provided a pictorial overview of the Feature Distribution Composition (FDC) Attention architecture and the attention can be utilized for different applications. FDC helps in deciding the features at a very high level of the whole image and mainly concentrates on determination of the combination of the lower level features instead of pre-likelihood or distribution. Since, we experimented different architectures and proposed different fusion models, we will discuss the utility from each prospect.  

\subsection{Multiple Role Representation Crossover Attention}
While, we utilized different features from the image for fusion in the network, like the global view for initialization and selection of weighted combination, we also introduce Multiple Role Representation Crossover (MRRC) Attention, where the aim is to provide attention, which is derived as a crossover of different aspect of the network. Residual and skip-through like architecture has been successful in many applications for gathering different residuals of the network. Some of our model do the same, where instead of utilization of the image features, it gathers the transformed features from some node of the network. Since, the depth of our MRRC architecture is very shallow, compared to ResNet, our residual collection can be any gate of the architecture, utilizing different aspects from images for generation and crossover the prospects to generate a new one with variations and developing the ability of the model to respond to those variations. 
Multiple Role Representation Crossover (MRRC) Attention Generation layer comprises of Image Weighted Tensors and are crossover-ed to generate the required attention dependent on time and a shifting weight scenario. The shifting of weights generate ample variations for the language decoder to be able to derive the sentences and generate diverse sentences. 
We can also call this attention as Image Weighted Tensor Crossover (IWTC) Attention when we use the FDC features as the prospects in place of $f(\textbf{I})$ in Equation \ref{eqn:MRRC}. 
Mathematically, Multiple Role Representation Crossover (MRRC) Attention Generation layer can be denoted by the following set of equations. 
\begin{equation} \label{eqn:MRRC}
\begin{split}
 \textbf{T}_{t} & =  
 \textbf{W}_{s_{12}}\text{ }\sigma(\textbf{W}_{s_{11}}\textbf{h}_{t-1} + \textbf{W}_{w_{_{1}}}\sum\limits_{i=0}^{t-1} \textbf{W}_e \textbf{x}_{i} + \textbf{b}_1) 
 \text{ } \otimes \\
 & \tanh (\textbf{W}_{s_{22}} ( f(\textbf{I}) \text{ }\sigma(\textbf{W}_{s_{21}}\textbf{h}_{t-1} + \textbf{W}_{w_{_{2}}}\sum\limits_{i=0}^{t-1} \textbf{W}_e \textbf{x}_{i} + \textbf{b}_2) ) + \textbf{b}_3) 
\end{split}
\end{equation}
where we have $\textbf{x} \in \mathbb{R}^{b \times d}$. 
\begin{figure}
\centering 
\includegraphics[]{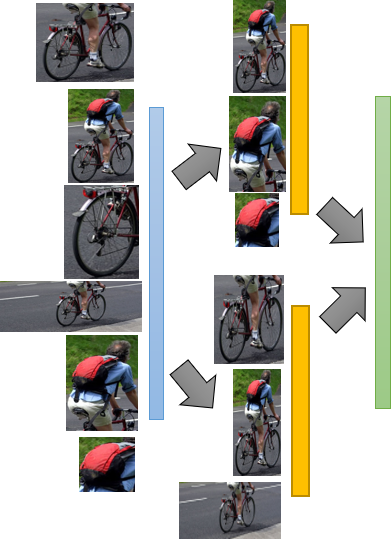}  
\caption{Image Weighted Tensor Crossover (IWTC) Generation Layer. Here, the regional image segments are decomosed to generate newer ones with better comprehension and information.}\label{fig:IWTC_generator}
\end{figure}
Figure \ref{fig:IWTC_generator} has given an overview of the Multiple Role Representation Crossover Attention architecture with more details of the involvement of the image characteristics and compositions. Multiple Role Representation Crossover Attention works on contexts related to the composition of the RCNN and thus can be considered as a dual prospect of FDC. In many situations, the weights of the neural network are biased to certain patterns and MRRC neutralizes the situation through extra weights and variances. This phenomenon is prevalent in statistics and is considered as interaction, where the variations are captured into the mathematical model through interaction term. However, due to the limitations in the number of considered variables in statistics, they are literally useless for intelligent systems. Hence, in this work, we have devised such kind of situation, where MRRC is derived from FDC and $\textbf{h}_{t-1}$. $\textbf{h}_{t-1}$ is the next state derivation from the RCNN features and must not confused as the hidden state of LSTM. In this work, we are talking about MRRC architecture, which does not comply with LSTM, as we never propagate the hidden state (like in LSTM). Instead, we deal with transformed RCNN features. Hence, this architecture is far different from the traditional recurrent neural network architectures. It is to be noted that MRRC also deals with the previous contexts and thus can be regarded as a much enhanced criteria for context generation and the topologically synchronized attention.

\subsection{Language Decoder}
Language Decoder operates alongside the attentions and previous contexts. Unlike LSTM, this language decoder operates on the influence of attentions and previous context and does not explicitly depend on hidden states for sequential relevance. The equations for Language Decoder is provided as the followings. 
\begin{equation}
 \textbf{i}_{t} = \sigma(\textbf{W}_{pi}\textbf{p}_{t} + \textbf{W}_{qi}\textbf{q}_{t} + \textbf{W}_{Ti}\textbf{T}_{t} + \textbf{b}_{i})
\end{equation}
\begin{equation}
 \textbf{f}_{t} = \sigma(\textbf{W}_{pf}\textbf{p}_{t} + \textbf{W}_{qf}\textbf{q}_{t} + \textbf{W}_{Tf}\textbf{T}_{t} + \textbf{b}_{f})
\end{equation}
\begin{equation}
 \textbf{o}_{t} = \sigma(\textbf{W}_{po}\textbf{p}_{t} + \textbf{W}_{qo}\textbf{q}_{t} + \textbf{W}_{To}\textbf{T}_{t} + \textbf{b}_{o})
\end{equation}
\begin{equation}
 \textbf{g}_{t} = \tanh(\textbf{W}_{pg}\textbf{p}_{t} + \textbf{W}_{qg}\textbf{q}_{t} + \textbf{W}_{Tg}\textbf{T}_{t} + \textbf{b}_{g})
\end{equation}
\begin{equation}
 \textbf{c}_{t} = \textbf{f}_{t} \odot \textbf{c}_{t-1} + \textbf{i}_{t} \odot \textbf{g}_{t} 
\end{equation}
\begin{equation}
 \textbf{h}_{t} = \textbf{o}_{t} \odot \tanh(\textbf{c}_{t})
\end{equation}
\begin{equation} \label{eq:lang_decode}
 \textbf{x}_{t} = \max \arg \mathrm{softmax} (\textbf{W}_{hx} \textbf{h}_{t})
\end{equation}
where the attentions $\textbf{q}_{t}$ and $\textbf{T}_{t}$ are derived from previous blocks and the $\textbf{p}_{t}$ is the previous transformed context embedding. 
While, most of the recurrent units are dependent on attentions ($\textbf{h}_{t-1}$, $\textbf{A}_{t}$) and pretext ($\textbf{W}_e\textbf{x}_{t}$), our language model depends on combinations, which are generated out of the local context from the feature space and what to select as local context is learned in the feature space. Next, we will discuss the variations of architectures of these concepts and how they are unique in generating the combinations of contexts. 



\subsection{FDC-MRRC}
Feature Distribution Composition - Multiple Role Representation Crossover (FDC-MRRC) is the basic architecture, where the main contributions are related to feature distribution generation and fusion of image weighted crossover attention. While FDC is about attention based on local enhancement, IWT provides a compositional construction that correlates with the local enhancement. This idea is motivated from the fact that both global and local prospect is important for diversified composition and establish local identity of feature space from the visual features. Figure \ref{fig:R5} provided a diagrammatic overview of the FDC-MRRC architecture and their individual components. Mathematically, we can define FDC-MRRC with the following set of equations. 
Starting with regional image features $\{v_1,\ldots,v_n\}$ for image $\textbf{I}$, $\overline{v} \in \mathbb{R}^{2048}$ is the average of all the possibilities over the feature space that the image represent.  
\begin{equation}
 \overline{v} = \frac{1}{k} \sum\limits_{i=1}^{i=k} v_i
\end{equation}
The initial parameters for the language decoder are initialized as the following for time $t=0$. 
\begin{equation}
 \textbf{h}_{0}, \textbf{ } \textbf{c}_{0} = \textbf{W}_{h_0}\overline{v}, \textbf{W}_{c_0}\overline{v}
\end{equation}
where the weights are defined as $\textbf{W}_{h_0} \in \mathbb{R}^{2048 \times d}$, $\textbf{W}_{h_0} \in \mathbb{R}^{2048 \times d}$ and these parameters effectively helps in determination of the  FDC for the language decoder. 
The intermediate FDC component is defined as $\widehat{v}_t$, through the help of the hidden components, which constitutes the status of previous composition and thus establishes the topological inference for the next instructions.  
\begin{equation} \label{eq:st1}
 \textbf{a}_{t} = \textbf{W}_{a} \tanh (\textbf{W}_{h} \textbf{h}_{t-1})
\end{equation}
where $\textbf{a}_{t} \in \mathbb{R}^{k \times 1}$, 
$\textbf{W}_{a} \in \mathbb{R}^{m_1 \times k}$, 
$\textbf{W}_{h} \in \mathbb{R}^{d \times m_1}$, 
$k$ is the number of considered regional features in images, $m_1$ is an intermediate dimension. Next, $\textbf{a}_{t}$ is reduced to a softmax layer and the new $\alpha_t$ is used for FDC composition. 
\begin{equation}
 \alpha_t = \mathrm{softmax}(\textbf{a}_t)
\end{equation}
with $\alpha_t = \{\alpha_{1,t},\ldots,\alpha_{k,t}\} \in \mathbb{R}^{k \times 1}$ and FDC composition is, 
\begin{equation} \label{eq:en1}
 \widehat{v}_t =  \sum\limits_{i=1}^{i=k} v_i \alpha_{i,t}
\end{equation}
where we have $\widehat{v}_t \in \mathbb{R}^{b \times 2048}$ where $b$ is the batch size and $d$ is the hidden layer dimension. We define new components $\textbf{q}_{t}$ and $\textbf{p}_{t}$ with $\textbf{W}_e \in \mathbb{R}^{V \times e}$ as the embedding vector for languages and is derived from Stanford GloVe and $V$ is the vocabulary size and $e$ is the dimension of the embedding.
\begin{equation}
 \textbf{q}_{t} = \widehat{v}_t
\end{equation}
\begin{equation}
 \textbf{p}_{t} = \textbf{W}_e \textbf{x}_{t-1}
\end{equation}
Next, we define the MRRC feature selection layer as $\textbf{T}_{t}$ as the following, 
\begin{equation}
\begin{split}
 \textbf{T}_{t} & =  
 \textbf{W}_{s_{12}}\text{ }\sigma(\textbf{W}_{s_{11}}f(\textbf{q}_{t}) + \textbf{W}_{w_{_{1}}}\sum\limits_{i=0}^{t-1} \textbf{W}_e \textbf{x}_{i} + \textbf{b}_1) 
 \text{ }\otimes \\
 & \tanh (\textbf{W}_{s_{22}} ( \textbf{q}_{t} \text{ }\sigma(\textbf{W}_{s_{21}}f(\textbf{q}_{t}) + \textbf{W}_{w_{_{2}}}\sum\limits_{i=0}^{t-1} \textbf{W}_e \textbf{x}_{i} + \textbf{b}_2) ) + \textbf{b}_3) \\
 & =  
 \textbf{W}_{s_{12}}\text{ }\sigma(\textbf{W}_{s_{11}}\textbf{h}_{t-1} + \textbf{W}_{w_{_{1}}}\sum\limits_{i=0}^{t-1} \textbf{W}_e \textbf{x}_{i} + \textbf{b}_1) 
 \text{ }\otimes \\
 & \tanh (\textbf{W}_{s_{22}} ( \textbf{q}_{t} \text{ }\sigma(\textbf{W}_{s_{21}}\textbf{h}_{t-1} + \textbf{W}_{w_{_{2}}}\sum\limits_{i=0}^{t-1} \textbf{W}_e \textbf{x}_{i} + \textbf{b}_2) ) + \textbf{b}_3) 
\end{split}
\end{equation}
where we have defined $\otimes$ as an algebraic operation (like tensor product or dot product or similar) for the matrices to generate a feature vector using the different composition states of the regional image features in $\textbf{h}_{t-1}$, which is more than hidden states of the language decoder. Here, we considered $\otimes = \odot$, element-wise multiplication, as we try to rectify and complement one context with the other context from the same image sources. Finally, we have,
\begin{equation}
 \textbf{i}_{t} = \sigma(\textbf{W}_{pi}\textbf{p}_{t} + \textbf{W}_{qi}\textbf{q}_{t} + \textbf{W}_{Ti}\textbf{T}_{t} + \textbf{b}_{i})
\end{equation}
\begin{equation}
 \textbf{f}_{t} = \sigma(\textbf{W}_{pf}\textbf{p}_{t} + \textbf{W}_{qf}\textbf{q}_{t} + \textbf{W}_{Tf}\textbf{T}_{t} + \textbf{b}_{f})
\end{equation}
\begin{equation}
 \textbf{o}_{t} = \sigma(\textbf{W}_{po}\textbf{p}_{t} + \textbf{W}_{qo}\textbf{q}_{t} + \textbf{W}_{To}\textbf{T}_{t} + \textbf{b}_{o})
\end{equation}
\begin{equation}
 \textbf{g}_{t} = \tanh(\textbf{W}_{pg}\textbf{p}_{t} + \textbf{W}_{qg}\textbf{q}_{t} + \textbf{W}_{Tg}\textbf{T}_{t} + \textbf{b}_{g})
\end{equation}
\begin{equation}
 \textbf{c}_{t} = \textbf{f}_{t} \odot \textbf{c}_{t-1} + \textbf{i}_{t} \odot \textbf{g}_{t} 
\end{equation}
\begin{equation}
 \textbf{h}_{t} = \textbf{o}_{t} \odot \tanh(\textbf{c}_{t})
\end{equation}
where we have $\textbf{h}_{t}$ as the pre-likelihood estimation of the topological sequence and is decoded as $\textbf{x}_t$ as in Equation \ref{eq:lang_decode}. 

Mathematically, Feature Distribution Composition - Multiple Role Representation Crossover (FDC-MRRC), denoted as $f_{_{FDC_1}}(.)$, can be described as the followings probability distribution estimation.
\begin{equation} \label{eq:itb1}
\begin{split}
 f & _{_{FDC_1}}(\textbf{I}) = \prod\limits_{k}^{} \mathrm{Pr}(w_k \mid \textbf{T}_t, \text{ } \Phi({v}_1. \ldots, {v}_K), \text{ } \textbf{W}_{L_1}) \\
 & \prod\limits_{i}^{} \mathrm{Pr}(\textbf{T}_t \mid f_1({v}_1. \ldots, {v}_K), \text{ }f_2({v}_1. \ldots, {v}_K),  \text{ }\textbf{W}_1)  \\
 & = \prod\limits_{k}^{} \mathrm{Pr}(w_k \mid \textbf{T}_t, \text{ }\Phi(\left( \frac{1}{K}\sum\limits_{m=1}^{K} {v}_m \right),  \left( \sum\limits_{m=1}^{N} v_m \alpha_{m}\right) , \textbf{W}_{L_1}  ) )  \\
 & \text{ }\text{ }\text{ } \prod\limits_{t}^{} \mathrm{Pr}(\textbf{T}_t \mid f_1({v}_1. \ldots, {v}_K), \text{ }f_2({v}_1. \ldots, {v}_K),  \text{ }\textbf{W}_1)  \\
 & = \prod\limits_{k}^{} \mathrm{Q}_{M}(w_k \mid \textbf{T}_t, \text{ }\Phi(\left( \frac{1}{K}\sum\limits_{m=1}^{K} {v}_m \right),  \left( \sum\limits_{m=1}^{N} v_m \alpha_{m}\right) ) )  \\
 & \text{ }\text{ }\text{ } \prod\limits_{t}^{} \mathrm{Q}_{F}(\textbf{T}_t \mid f_1({v}_1. \ldots, {v}_K), \text{ }f_2({v}_1. \ldots, {v}_K))  
\end{split} 
\end{equation}
using the weights of the LSTM in the architecture is denoted as $\textbf{W}_{L_1}$, $w_i$ as words of sentences, $v_i$ as regional image features, ${a}_iv_m$ as intermediate learnt parameters, $\mathrm{Q}_M(.)$ and $\mathrm{Q}_F(.)$ are the Image Caption generator and feature generator function respectively. $\mathrm{Q}_F(.)$ derives $\textbf{T}_i$ from $\textbf{v}$ of $\textbf{I}$.

\begin{figure*}
\centering 
\includegraphics[width=\textwidth]{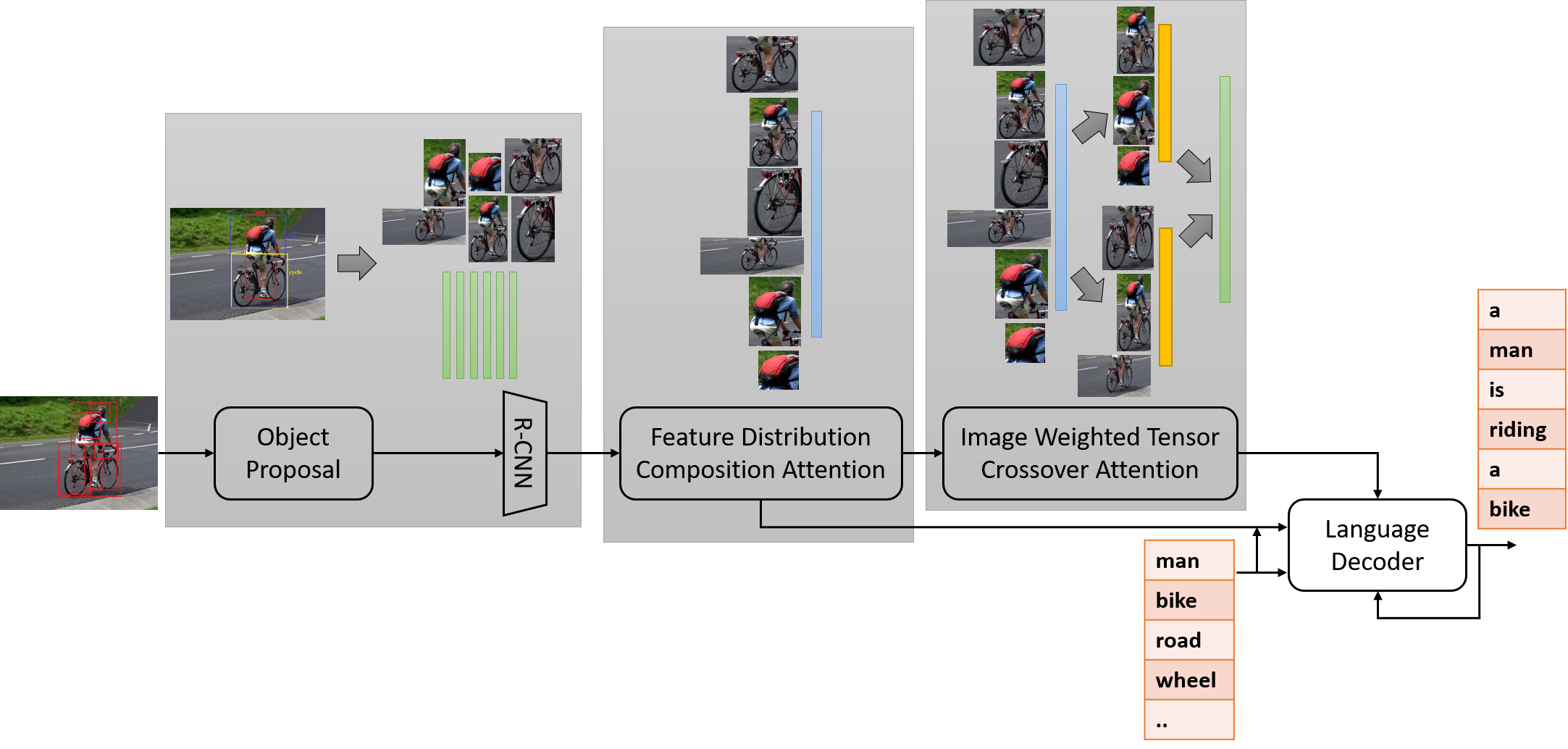}  
\caption{Overall Semi-Factorized-FDC-MRRC Architecture With FDC, IWTC, Semantic and RCNN Feature Fusion.}
\label{fig:R5_}
\end{figure*}

\subsection{Semi-Factorized-FDC-MRRC}
Semi-Factorized Feature Distribution Composition - Multiple Role Representation Crossover (Semi-Factorized-FDC-MRRC) creates a factorized version of the $\textbf{q}_{t}$ before being decoded for the topological interpretation. 
Figure \ref{fig:R5_} has provided a pictorial overview of the Semi-Factorized-FDC-MRRC  architecture. From the prospect of feature composition, this is much more robust as these are dealing with factorization of the feature space to create diversity. Also, the FDC features are derived from regional local importance of the images based on RPN (Region Prediction Network), whose works is to predict the regions of importance based on previously accounted probable regions. While, factorization for neural space is not equivalent to matrix factorization based on heuristics. This is because of the fact that such factorization do not scale well for applications and difficult to find the optimal space of factors. Also, for decoding tasks, such factorized model may not be helpful and do not fit well. Hence, a weighted factorization is introduced which helps in determination of partial usefulness of the feature vector that can be utilized for the model.  
Mathematically, we can define Semi-Factorized-FDC-MRRC as the following equations, where  most of the notations are described for FDC-MRRC architecture. 
\begin{equation}
 \overline{v} = \frac{1}{k} \sum\limits_{i=1}^{i=k} v_i
\end{equation}
The initial parameters are initialized as, 
\begin{equation}
 \textbf{h}_{0}, \textbf{ } \textbf{c}_{0} = \textbf{W}_{h_0}\overline{v}, \textbf{W}_{c_0}\overline{v}
\end{equation}
\begin{equation} \label{eq:st2}
 \textbf{a}_{t} = \textbf{W}_{a} \tanh (\textbf{W}_{h} \textbf{h}_{t-1})
\end{equation}
\begin{equation}
 \alpha_t = \mathrm{softmax}(\textbf{a}_t)
\end{equation}
\begin{equation} \label{eq:en2}
 \widehat{v}_t =  \sum\limits_{i=1}^{i=k} v_i \alpha_{i,t}
\end{equation}
The assembling and selector layer is defined for IWT as, 
\begin{equation}
 \textbf{q}_{t} = \widehat{v}_t
\end{equation}
\begin{equation}
 \textbf{p}_{t} = \textbf{W}_e \textbf{x}_{t-1}
\end{equation}
\begin{equation}
\begin{split}
 \textbf{T}_{t} & =  
 \textbf{W}_{s_{12}}\text{ }\sigma(\textbf{W}_{s_{11}}f(\textbf{q}_{t}) + \textbf{W}_{w_{_{1}}}\sum\limits_{i=0}^{t-1} \textbf{W}_e \textbf{x}_{i} + \textbf{b}_1) 
 \text{ }\otimes \\
 & \tanh (\textbf{W}_{s_{22}} ( \textbf{q}_{t} \text{ }\sigma(\textbf{W}_{s_{21}}f(\textbf{q}_{t}) + \textbf{W}_{w_{_{2}}}\sum\limits_{i=0}^{t-1} \textbf{W}_e \textbf{x}_{i} + \textbf{b}_2) ) + \textbf{b}_3) \\
 & =  
 \textbf{W}_{s_{12}}\text{ }\sigma(\textbf{W}_{s_{11}}\textbf{h}_{t-1} + \textbf{W}_{w_{_{1}}}\sum\limits_{i=0}^{t-1} \textbf{W}_e \textbf{x}_{i} + \textbf{b}_1) 
 \text{ }\otimes \\
 & \tanh (\textbf{W}_{s_{22}} ( \textbf{q}_{t} \text{ }\sigma(\textbf{W}_{s_{21}}\textbf{h}_{t-1} + \textbf{W}_{w_{_{2}}}\sum\limits_{i=0}^{t-1} \textbf{W}_e \textbf{x}_{i} + \textbf{b}_2) ) + \textbf{b}_3) 
\end{split}
\end{equation}
The next equation introduces the factorization portion for the FDC $\textbf{q}_{t}$ and is feed as $\textbf{q}_{t,n}$ in place of each $\textbf{q}_{t}$ for $\textbf{i}_{t}$, $\textbf{f}_{t}$, $\textbf{o}_{t}$ and $\textbf{g}_{t}$ of the language decoder. 
Equation \ref{eq:tag2} is called factorization step as the shared weights are decomposed into relevant slices of structured components with the help of a pre-trained distribution of the objects in the images. The main reason of factorization is to keep track of the composition of objects in images. Normally, FDC composition from RCNN objects feature representation will be heuristic and depend on the RPN prediction of Faster-RCNN, hence, factorization can help in establishing the structural properties between RCNN features and distribution of objects much faster. Unlike, previous works that operate on weights, directly related to the image features, our architecture is dependent on FDC composition and the factorized weights is expected to know the FDC composition than the whole image. 

Mathematically, Semi-Factorized Feature Distribution Composition - Multiple Role Representation Crossover (Semi-Factorized-FDC-MRRC), denoted as $f_{_{FDC_2}}(.)$, can be described as the followings probability distribution estimation.
\begin{equation} \label{eq:itb2}
\begin{split}
 f & _{_{FDC_2}}(\textbf{I}) = \prod\limits_{k}^{} \mathrm{Pr}(w_k \mid \textbf{T}_t, \text{ } \Phi({v}_1. \ldots, {v}_K), \text{ }S, \text{ } \textbf{W}_{L_1}) \\
 & \prod\limits_{t}^{} \mathrm{Pr}(\textbf{T}_t \mid f_1({v}_1. \ldots, {v}_K), \text{ }f_2({v}_1. \ldots, {v}_K),  \text{ }\textbf{W}_1)  \\
 & = \prod\limits_{k}^{} \mathrm{Pr}(w_k \mid \textbf{T}_t, \text{ }\Phi(\left( \frac{1}{K}\sum\limits_{m=1}^{K} {v}_m \right),  \left( \sum\limits_{m=1}^{N} v_m \alpha_{m}\right) , \text{ }S)   \\
 & , \text{ }\textbf{W}_{L_1}  )      \text{ }\text{ }\text{ } \prod\limits_{t}^{} \mathrm{Pr}(\textbf{T}_t \mid f_1({v}_1. \ldots, {v}_K), \text{ }f_2({v}_1. \ldots, {v}_K),  \text{ }\textbf{W}_1)  \\
 & = \prod\limits_{k}^{} \mathrm{Q}_{M}(w_k \mid \textbf{T}_t, \text{ }\Phi(\left( \frac{1}{K}\sum\limits_{m=1}^{K} {v}_m \right),  \left( \sum\limits_{m=1}^{N} v_m \alpha_{m}\right), \text{ }S ) )  \\
 & \text{ }\text{ }\text{ } \prod\limits_{t}^{} \mathrm{Q}_{F}(\textbf{T}_t \mid f_1({v}_1. \ldots, {v}_K), \text{ }f_2({v}_1. \ldots, {v}_K))  
\end{split} 
\end{equation}
using the weights of the LSTM in the architecture is denoted as $\textbf{W}_{L_1}$, $w_i$ as words of sentences, $v_i$ as regional image features, ${a}_iv_m$ as intermediate learnt parameters, $\mathrm{Q}_M(.)$ and $\mathrm{Q}_F(.)$ are the Image Caption generator and feature generator function respectively. $\mathrm{Q}_F(.)$ derives $\textbf{T}_i$ from $\textbf{v}$ of $\textbf{I}$, $S$ is the semantic feature map.

\subsubsection{Pseudo-Weights}
The tags are pseudo-weights that are not learned but derived from a model and is expected to behave like weights. As an adaptive weight concept, these make different impacts on the feature composition. These are also called tag based weighted metrics and form a series of tensors working with each other. We can have any number of such tensors (parameters), however, for prove of concept and feasibility of learning, we have considered three matrix product setup, reducing the learnable weights with high performing model derivatives. We consider  \textbf{W} decomposition is given by, 
\begin{equation}
\begin{split}
 \textbf{W} & =  \textbf{W}_{p_r} \text{ } \text{diag}( \textbf{W}_q \textbf{q}) \text{ } \textbf{W}_{p_s}
  \\
 & =  \textbf{W}_{p_r} \text{ } \text{diag}( \textbf{W}_q \Phi(v_1,v_2,\ldots,v_k)) \text{ } \textbf{W}_{p_s}
\end{split}
\end{equation} 
where $\textbf{W}_{p_r}$ and $\textbf{W}_{p_s}$ are two matrices accessed by all possible training scenarios and captions, while $\text{diag}( \textbf{W}_q \textbf{q})$ or $\text{diag}( \textbf{W}_q \Phi(v_1,v_2,\ldots,v_k))$ is a model derived transformation of the scenarios vectors. $\text{diag}( \textbf{W}_q \textbf{q})$ or $\text{diag}( \textbf{W}_q \Phi(v_1,v_2,\ldots,v_k))$ also represents common linguistic topological features to establish the grammar in sentences and encourage appearance of adverb and adjective level complexity.
$diag(.)$, accounts for $k$ semantic aspects in the form of orthogonal spaces to capture the required features. Each slice of the weight tensor (diagonal matrices) corresponds to a concept and varies in size with the dataset.

\subsubsection{Dynamic Factorization}
In this work, we will discuss Dynamic Factorization, as this has the potential of demonstrating much better variation in representation for caption generation. Dynamic Factorization helps in better and optimized decomposition of the lower level components (regional features of the images) and creates the chances of rectification in case of low usefulness of the factorization context vectors (semantics) from images.
\begin{figure*}
\centering 
\includegraphics[width=0.95\textwidth]{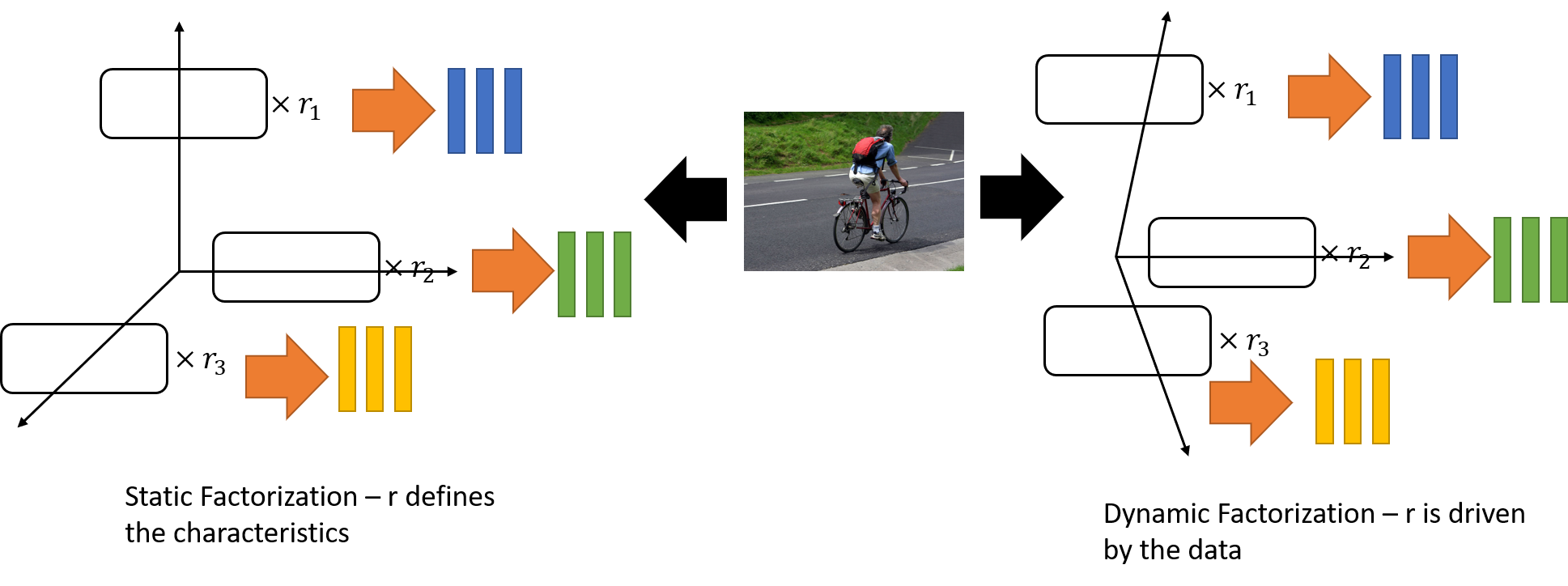}  
\caption{Dynamic Factorization and Static Factorization}\label{fig:Factorization}
\end{figure*}
Figure \ref{fig:Factorization} provide a diagram differentiating Dynamic Factorization with Static Factorization. Mathematically, we can define this new factorization approach (Dynamic Factorization) as the following,
\begin{equation} \label{eq:dyna}
\begin{split}
 \textbf{W} & =  \textbf{W}_{p_r} \text{ } \text{diag}( \textbf{W}_q \textbf{q}) \text{ } \textbf{W}_{p_s}
  \\
 & =  \textbf{W}_{p_r} \text{ } \text{diag}( \textbf{W}_q \Phi(v_1,v_2,\ldots,v_k)) \text{ } \textbf{W}_{p_s}
\end{split}
\end{equation}
Compared to the previous factorization approach (Static Factorization), which were defined as,  
\begin{equation} \label{eq:stat}
\begin{split}
 \textbf{W} & =  \textbf{W}_{p_r} \text{ } \text{diag}( \textbf{W}_q \textbf{q}) \text{ } \textbf{W}_{p_s}
  \\
 & = \textbf{W}_{p_r} \text{ } \text{diag}( \textbf{W}_q \Phi(\textbf{v})) \text{ } \textbf{W}_{p_s}
 \\
 & = \textbf{W}_{p_r} \text{ } \text{diag}( \textbf{W}_q S) \text{ } \textbf{W}_{p_s} 
\end{split}
\end{equation}
where $\Phi(.)$ is a functional transformation.
Finally, we define the equation for the new $\textbf{q}_{t,n}$ as,
\begin{equation} \label{eq:tag2}
\textbf{q}_{t,n} = \textbf{W}_{q,m} S \odot \textbf{W}_{q,n} \textbf{q}_{t}
\end{equation} 
where $\textbf{q}_{t,n} \in \mathbb{R}^{b \times d}$,
$\textbf{q}_{t} \in \mathbb{R}^{b \times 999}$,
$S \in \mathbb{R}^{b \times 999}$,
$\textbf{W}_{q,m} \in \mathbb{R}^{999 \times d}$,
$\textbf{W}_{q,n} \in \mathbb{R}^{2048 \times d}$.
The final Language Decoder is defined as the followings and in place of $\textbf{W}_{q*} \in \mathbb{R}^{2048 \times d}$, we have $\textbf{W}_{q*} \in \mathbb{R}^{d \times d}$ for $* = i/f/o/g$ and for some value of dimension $m_2$. 
\begin{equation}
 \textbf{i}_{t} = \sigma(\textbf{W}_{pi}\textbf{p}_{t} + \textbf{W}_{qi}\textbf{q}_{t,n} + \textbf{W}_{Ti}\textbf{T}_{t} + \textbf{b}_{i})
\end{equation}
\begin{equation}
 \textbf{f}_{t} = \sigma(\textbf{W}_{pf}\textbf{p}_{t} + \textbf{W}_{qf}\textbf{q}_{t,n} + \textbf{W}_{Tf}\textbf{T}_{t} + \textbf{b}_{f})
\end{equation}
\begin{equation}
 \textbf{o}_{t} = \sigma(\textbf{W}_{po}\textbf{p}_{t} + \textbf{W}_{qo}\textbf{q}_{t,n} + \textbf{W}_{To}\textbf{T}_{t} + \textbf{b}_{o})
\end{equation}
\begin{equation}
 \textbf{g}_{t} = \tanh(\textbf{W}_{pg}\textbf{p}_{t} + \textbf{W}_{qg}\textbf{q}_{t,n} + \textbf{W}_{Tg}\textbf{T}_{t} + \textbf{b}_{g})
\end{equation}
\begin{equation}
 \textbf{c}_{t} = \textbf{f}_{t} \odot \textbf{c}_{t-1} + \textbf{i}_{t} \odot \textbf{g}_{t} 
\end{equation}
\begin{equation}
 \textbf{h}_{t} = \textbf{o}_{t} \odot \tanh(\textbf{c}_{t})
\end{equation}
where $\textbf{h}_{t}$ is pre-likelihood estimation of the topological sequence and is decoded as $\textbf{x}_t$ as in Equation \ref{eq:lang_decode}.

\begin{figure*}
\centering 
\includegraphics[width=\textwidth]{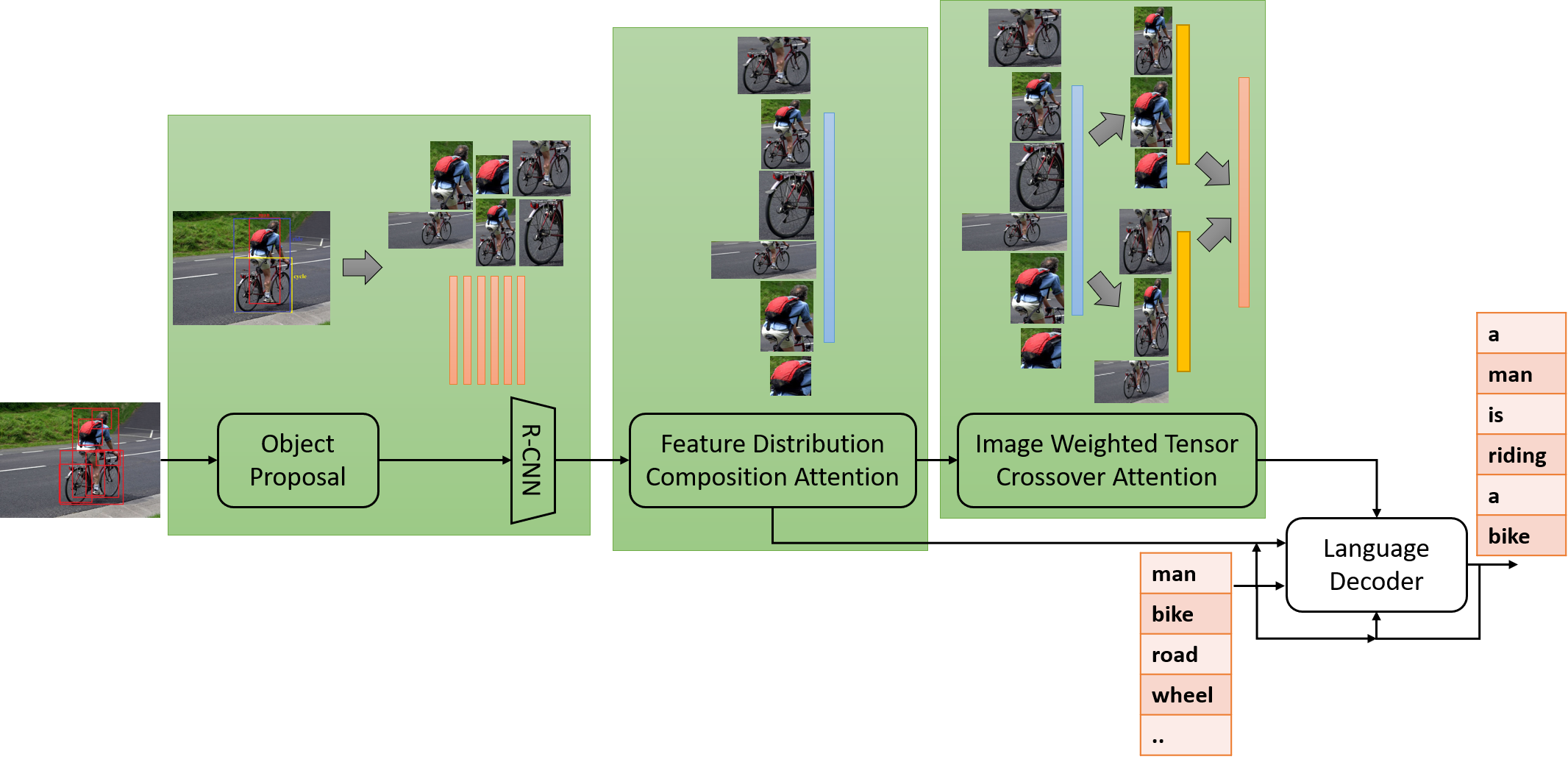}  
\caption{Overall Full-Factorized-FDC-MRRC Architecture With FDC, IWTC, Semantic and RCNN Feature Fusion.}
\label{fig:R6}
\end{figure*}

\subsection{Full-Factorized-FDC-MRRC}
Full-Factorized Feature Distribution Composition - Multiple Role Representation Crossover (Full-Factorized-FDC-MRRC) uses more factorized components for the language decoders for better enhancement of the features for the language decoder, where apart from decomposition of the weights for the image features FDC, the previous context embedding is also disintegrated. This disintegration of the language embedding brings stability in representation, which are unable to capture (or approximated) by weights. Figure \ref{fig:R6} provided the Full-Factorized-FDC-MRRC architecture details in a diagrammatic form for better perception of the network. The equations for the mathematical details of Full-Factorized-FDC-MRRC is denoted as the followings, 
\begin{equation}
 \overline{v} = \frac{1}{k} \sum\limits_{i=1}^{i=k} v_i
\end{equation} 
\begin{equation}
 \textbf{h}_{0}, \textbf{ } \textbf{c}_{0} = \textbf{W}_{h_0}\overline{v}, \textbf{W}_{c_0}\overline{v}
\end{equation}
\begin{equation} \label{eq:st3}
 \textbf{a}_{t} = \textbf{W}_{a} \tanh (\textbf{W}_{h} \textbf{h}_{t-1})
\end{equation}
\begin{equation}
 \alpha_t = \mathrm{softmax}(\textbf{a}_t)
\end{equation}
\begin{equation} \label{eq:en3}
 \widehat{v}_t =  \sum\limits_{i=1}^{i=k} v_i \alpha_{i,t}
\end{equation}
\begin{equation}
 \textbf{q}_{t} = \widehat{v}_t
\end{equation}
\begin{equation}
 \textbf{p}_{t} = \textbf{W}_e \textbf{x}_{t-1}
\end{equation}
\begin{equation}
\begin{split}
 \textbf{T}_{t} & =  
 \textbf{W}_{s_{12}}\text{ }\sigma(\textbf{W}_{s_{11}}f(\textbf{q}_{t}) + \textbf{W}_{w_{_{1}}}\sum\limits_{i=0}^{t-1} \textbf{W}_e \textbf{x}_{i} + \textbf{b}_1) 
 \text{ }\otimes \\
 & \tanh (\textbf{W}_{s_{22}} ( \textbf{q}_{t} \text{ }\sigma(\textbf{W}_{s_{21}}f(\textbf{q}_{t}) + \textbf{W}_{w_{_{2}}}\sum\limits_{i=0}^{t-1} \textbf{W}_e \textbf{x}_{i} + \textbf{b}_2) ) + \textbf{b}_3) \\
 & =  
 \textbf{W}_{s_{12}}\text{ }\sigma(\textbf{W}_{s_{11}}\textbf{h}_{t-1} + \textbf{W}_{w_{_{1}}}\sum\limits_{i=0}^{t-1} \textbf{W}_e \textbf{x}_{i} + \textbf{b}_1) 
 \text{ }\otimes \\
 & \tanh (\textbf{W}_{s_{22}} ( \textbf{q}_{t} \text{ }\sigma(\textbf{W}_{s_{21}}\textbf{h}_{t-1} + \textbf{W}_{w_{_{2}}}\sum\limits_{i=0}^{t-1} \textbf{W}_e \textbf{x}_{i} + \textbf{b}_2) ) + \textbf{b}_3) 
\end{split}
\end{equation}
For every $* = i/f/o/g$, we define separate factorization of the different language components and can be defined as the followings.  
\begin{equation}
 \textbf{p}_{*,t} = \textbf{W}_{p,*m} S \odot \textbf{W}_{p,*n} \textbf{p}_{t}
\end{equation}
\begin{equation}
 \textbf{q}_{*,t} = \textbf{W}_{q,*m} S \odot \textbf{W}_{q,*n} \textbf{q}_{t}
\end{equation}
where $\textbf{W}_{p,*m} \in \mathbb{R}^{999 \times d}$,
$\textbf{W}_{p,*n} \in \mathbb{R}^{e \times d}$, $e$ is the dimension of the embedding for $\textbf{W}_e \in \mathbb{R}^{V \times e}$, $V$ is the vocabulary size, 
$\textbf{W}_{q,*m} \in \mathbb{R}^{999 \times d}$,
$\textbf{W}_{q,*n} \in \mathbb{R}^{2048 \times d}$. 
It is a definite practice in neural network to factorize the language components. This is because of their concrete structures in feature space, acquired through transfer learning like word embedding and language distribution from images. However, these factorized weights help in better topic modeling and understanding the required direction of operation instead of estimating the representation into a definite direction through non-linear transformation based estimations of distribution. 
Unlike previous language decoders, we have $\textbf{W}_{p*} \in \mathbb{R}^{d \times d}$ and $\textbf{W}_{q*} \in \mathbb{R}^{d \times d}$ for the following equations. 
\begin{equation}
 \textbf{i}_{t} = \sigma(\textbf{W}_{pi}\textbf{p}_{i,t} + \textbf{W}_{qi}\textbf{q}_{i,t} + \textbf{W}_{Ti}\textbf{T}_{t} + \textbf{b}_{i})
\end{equation}
\begin{equation}
 \textbf{f}_{t} = \sigma(\textbf{W}_{pf}\textbf{p}_{f,t} + \textbf{W}_{qf}\textbf{q}_{f,t} + \textbf{W}_{Tf}\textbf{T}_{t} + \textbf{b}_{f})
\end{equation}
\begin{equation}
 \textbf{o}_{t} = \sigma(\textbf{W}_{po}\textbf{p}_{o,t} + \textbf{W}_{qo}\textbf{q}_{o,t} + \textbf{W}_{To}\textbf{T}_{t} + \textbf{b}_{o})
\end{equation}
\begin{equation}
 \textbf{g}_{t} = \tanh(\textbf{W}_{pg}\textbf{p}_{g,t} + \textbf{W}_{qg}\textbf{q}_{g,t} + \textbf{W}_{Tg}\textbf{T}_{t} + \textbf{b}_{g})
\end{equation}
\begin{equation}
 \textbf{c}_{t} = \textbf{f}_{t} \odot \textbf{c}_{t-1} + \textbf{i}_{t} \odot \textbf{g}_{t} 
\end{equation}
\begin{equation}
 \textbf{h}_{t} = \textbf{o}_{t} \odot \tanh(\textbf{c}_{t})
\end{equation}
where $\textbf{h}_{t}$ is pre-likelihood estimation of the topological sequence and is decoded as $\textbf{x}_t$ as in Equation \ref{eq:lang_decode}.

Mathematically, Full-Factorized Feature Distribution Composition - Multiple Role Representation Crossover (Full-Factorized-FDC-MRRC), denoted as $f_{_{FDC_3}}(.)$, can be described as the followings probability distribution estimation.
\begin{equation} \label{eq:itb3}
\begin{split}
 f & _{_{FDC_3}}(\textbf{I}) = \prod\limits_{k}^{} \mathrm{Pr}(w_k \mid \textbf{T}_t, \text{ } \Phi({v}_1. \ldots, {v}_K), \text{ }S, \text{ } f_L(\textbf{W}_{L_1}, \text{ }S)) \\
 & \prod\limits_{i}^{} \mathrm{Pr}(\textbf{T}_t \mid f_1({v}_1. \ldots, {v}_K), \text{ }f_2({v}_1. \ldots, {v}_K),  \text{ }\textbf{W}_1)  \\
 & = \prod\limits_{k}^{} \mathrm{Pr}(w_k \mid \textbf{T}_t, \text{ }\Phi(\left( \frac{1}{K}\sum\limits_{m=1}^{K} {v}_m \right),  \left( \sum\limits_{m=1}^{N} v_m \alpha_{m}\right) , \text{ }S  ),  \\
 & \text{ }f_L(\textbf{W}_{L_1}, \text{ }S)  )   \text{ }\text{ }\text{ } \prod\limits_{t}^{} \mathrm{Pr}(\textbf{T}_t \mid f_1({v}_1. \ldots, {v}_K), \text{ }f_2({v}_1. \ldots, {v}_K),  \text{ }\textbf{W}_1)  \\
 & = \prod\limits_{k}^{} \mathrm{Q}_{M}(w_k \mid \textbf{T}_t, \text{ }\Phi(\left( \frac{1}{K}\sum\limits_{m=1}^{K} {v}_m \right),  \left( \sum\limits_{m=1}^{N} v_m \alpha_{m}\right), \text{ }S ) )  \\
 & \text{ }\text{ }\text{ } \prod\limits_{t}^{} \mathrm{Q}_{F}(\textbf{T}_t \mid f_1({v}_1. \ldots, {v}_K), \text{ }f_2({v}_1. \ldots, {v}_K))  
\end{split} 
\end{equation}
using the weights of the LSTM in the architecture is denoted as $\textbf{W}_{L_1}$, $w_i$ as words of sentences, $v_i$ as regional image features, ${a}_iv_m$ as intermediate learnt parameters, $\mathrm{Q}_M(.)$ and $\mathrm{Q}_F(.)$ are the Image Caption generator and feature generator function respectively. $\mathrm{Q}_F(.)$ derives $\textbf{T}_i$ from $\textbf{v}$ of $\textbf{I}$, $S$ is the semantic feature map.

\begin{figure*}
\centering 
\includegraphics[width=\textwidth]{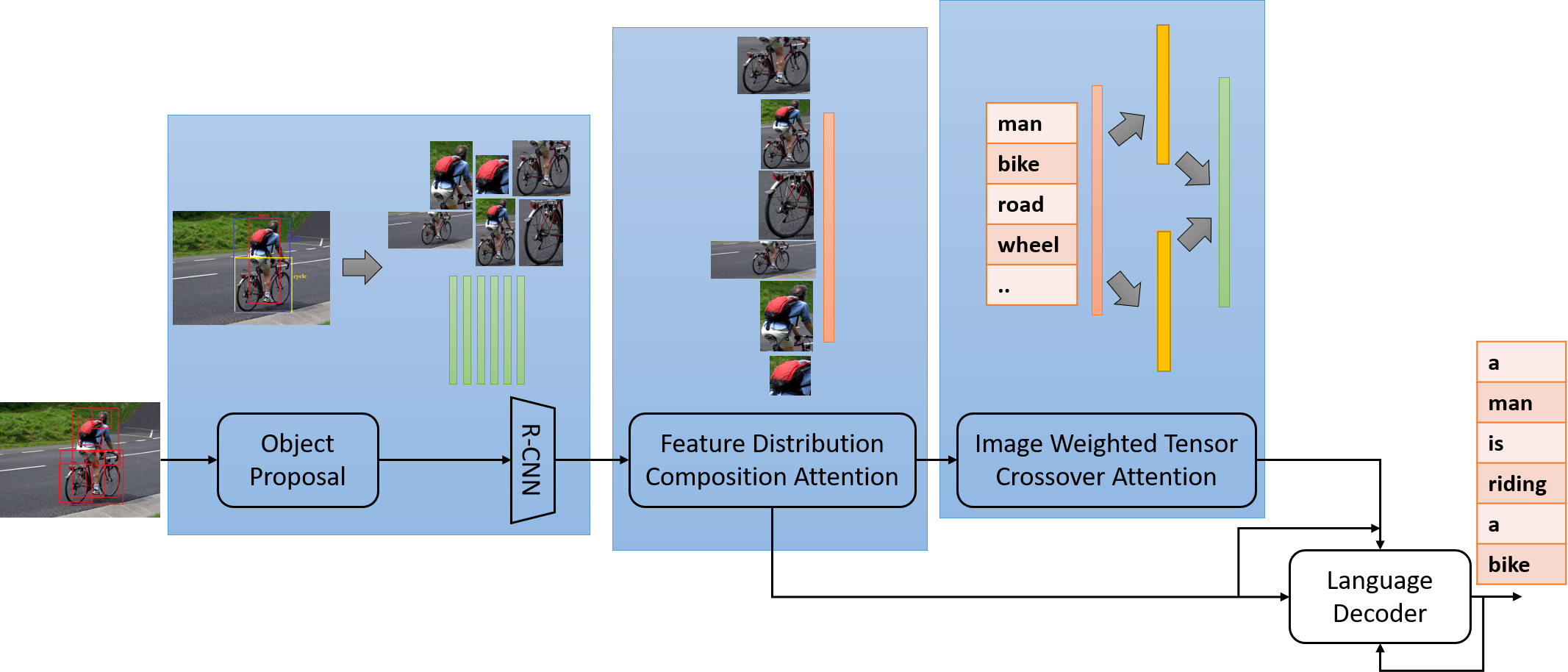}  
\caption{Overall Semi-FDC-Factorized-Semantic-MRRC Architecture With FDC, IWTC, Semantic and RCNN Feature Fusion.}
\label{fig:fdc}
\end{figure*}

\subsection{Semi-FDC-Factorized-Semantic-MRRC}
Semi Feature Distribution Composition Factorized - Semantic - Multiple Role Representation Crossover (Semi-FDC-Factorized-Semantic-MRRC) operates on more refinement of the feature space decomposition through the utilization of semantic distribution for IWT Crossover Attention generation, which was previously dependent on regional image features. Regional image features are generated on the prediction of RPN and its position in the feature vector $\{v_1,v_2,\ldots,v_k\}$ is random as there is no criteria to establish the correctness of the sequence. However, since the features are extracted on a fine tuned model, it is expected that the model learned to establish the relative topological usefulness. Regional image features have this drawback and when we counter it through semantic based IWT Crossover attention, there ws enhancement in performance. 
Figure \ref{fig:fdc} provided the figure for Semi-FDC-Factorized-Semantic-MRRC architecture in details. Mathematical equations that govern the Semi-FDC-Factorized-Semantic-MRRC architecture can be denoted as the followings and most of the dimensions and notations are similar as before unless stated. 
\begin{equation}
 \overline{v} = \frac{1}{k} \sum\limits_{i=1}^{i=k} v_i
\end{equation} 
\begin{equation}
 \textbf{h}_{0}, \textbf{ } \textbf{c}_{0} = \textbf{W}_{h_0}\overline{v}, \textbf{W}_{c_0}\overline{v}
\end{equation}
\begin{equation} \label{eq:st4}
 \textbf{a}_{t} = \textbf{W}_{a} \tanh (\textbf{W}_{h} \textbf{h}_{t-1})
\end{equation}
\begin{equation}
 \alpha_t = \mathrm{softmax}(\textbf{a}_t)
\end{equation}
\begin{equation} \label{eq:en4}
 \textbf{q}_{t} = \widehat{v}_t =  \sum\limits_{i=1}^{i=k} v_i \alpha_{i,t}
\end{equation}
\begin{equation}
 \textbf{p}_{t} = \textbf{W}_e \textbf{x}_{t-1}
\end{equation}
The MRRC equation has been replaced with a function of semantic level representation $S$ as $f(S)$ has replaced $f(\textbf{q}_{t})$ to generate the MRRC attention. While, most of the factorization of vectors were helping in characterization, we introduce MRRC attention as characterized attention and is defined as the following equation.  
\begin{equation}
\begin{split}
 \textbf{T}_{t} & =  
 \textbf{W}_{s_{12}}\text{ }\sigma(\textbf{W}_{s_{11}}f(S) + \textbf{W}_{w_{_{1}}}\sum\limits_{i=0}^{t-1} \textbf{W}_e \textbf{x}_{i} + \textbf{b}_1) 
 \text{ }\otimes \\
 & \tanh (\textbf{W}_{s_{22}} ( S \text{ }\sigma(\textbf{W}_{s_{21}}f(\textbf{q}_{t}) + \textbf{W}_{w_{_{2}}}\sum\limits_{i=0}^{t-1} \textbf{W}_e \textbf{x}_{i} + \textbf{b}_2) ) + \textbf{b}_3) \\
 & =  
 \textbf{W}_{s_{12}}\text{ }\sigma(\textbf{W}_{s_{11}}S + \textbf{W}_{w_{_{1}}}\sum\limits_{i=0}^{t-1} \textbf{W}_e \textbf{x}_{i} + \textbf{b}_1) 
 \text{ }\otimes \\
 & \tanh (\textbf{W}_{s_{22}} ( S \text{ }\sigma(\textbf{W}_{s_{21}}\textbf{h}_{t-1} + \textbf{W}_{w_{_{2}}}\sum\limits_{i=0}^{t-1} \textbf{W}_e \textbf{x}_{i} + \textbf{b}_2) ) + \textbf{b}_3) 
\end{split}
\end{equation}
The Language Decoder follows similar strategy as Semi-Factorized-FDC-MRRC, but the factorization is aided by the $\textbf{q}_{t}$ instead of  $S$. From this approach, we introduce the notion of a Dynamic Factorization scheme instead of Static Factorization. 
While Static Factorization is defined as Equation \ref{eq:stat}, Dynamic Factorization is defined as Equation \ref{eq:dyna}. Here, we have used Dynamic Factorization where the factorized component changes with time and context and performed much better than previous models. 
\begin{equation} \label{eq:tag4}
\textbf{q}_{t,n} = \textbf{W}_{h,m} f(\textbf{q}_{t-1}) \odot \textbf{W}_{h,n} \textbf{q}_{t} = \textbf{W}_{h,m} \textbf{h}_{t-1} \odot \textbf{W}_{h,n} \textbf{q}_{t}
\end{equation} 
\begin{equation}
 \textbf{i}_{t} = \sigma(\textbf{W}_{pi}\textbf{p}_{t} + \textbf{W}_{qi}\textbf{q}_{t,n} + \textbf{W}_{Ti}\textbf{T}_{t} + \textbf{b}_{i})
\end{equation}
\begin{equation}
 \textbf{f}_{t} = \sigma(\textbf{W}_{pf}\textbf{p}_{t} + \textbf{W}_{qf}\textbf{q}_{t,n} + \textbf{W}_{Tf}\textbf{T}_{t} + \textbf{b}_{f})
\end{equation}
\begin{equation}
 \textbf{o}_{t} = \sigma(\textbf{W}_{po}\textbf{p}_{t} + \textbf{W}_{qo}\textbf{q}_{t,n} + \textbf{W}_{To}\textbf{T}_{t} + \textbf{b}_{o})
\end{equation}
\begin{equation}
 \textbf{g}_{t} = \tanh(\textbf{W}_{pg}\textbf{p}_{t} + \textbf{W}_{qg}\textbf{q}_{t,n} + \textbf{W}_{Tg}\textbf{T}_{t} + \textbf{b}_{g})
\end{equation}
\begin{equation}
 \textbf{c}_{t} = \textbf{f}_{t} \odot \textbf{c}_{t-1} + \textbf{i}_{t} \odot \textbf{g}_{t} 
\end{equation}
\begin{equation}
 \textbf{h}_{t} = \textbf{o}_{t} \odot \tanh(\textbf{c}_{t})
\end{equation}
where $\textbf{h}_{t}$ is pre-likelihood estimation of the topological sequence and is decoded as $\textbf{x}_t$ as in Equation \ref{eq:lang_decode}.

Mathematically, Semi Feature Distribution Composition Factorized - Semantic - Multiple Role Representation Crossover (Semi-FDC-Factorized-Semantic-MRRC), denoted as $f_{_{FDC_4}}(.)$, can be described as the followings probability distribution estimation.
\begin{equation} \label{eq:itb4}
\begin{split}
 f & _{_{FDC_4}}(\textbf{I}) = \prod\limits_{k}^{} \mathrm{Pr}(w_k \mid \textbf{T}_t, \text{ } \Phi({v}_1. \ldots, {v}_K), \text{ }S, \text{ } \textbf{W}_{L_1}) \\
 & \prod\limits_{t}^{} \mathrm{Pr}(\textbf{T}_t \mid f_1({v}_1. \ldots, {v}_K, S),  \text{ }f_2({v}_1. \ldots, {v}_K, S), \text{ }\textbf{W}_1)  \\
 & = \prod\limits_{k}^{} \mathrm{Pr}(w_k \mid \textbf{T}_t, \text{ }\Phi(\left( \frac{1}{K}\sum\limits_{m=1}^{K} {v}_m \right), \\
 &  \left( \textbf{W}_x \textbf{h}_{t-1} \sum\limits_{m=1}^{N} v_m \alpha_{m} \text{ } \textbf{W}_y \right) , \text{ }S ), \text{ } \textbf{W}_{L_1}  )  \\
 &      \text{ }\text{ }\text{ } \prod\limits_{t}^{} \mathrm{Pr}(\textbf{T}_t \mid f_1({v}_1. \ldots, {v}_K, S), \text{ } f_2({v}_1. \ldots, {v}_K, S),  \text{ }\textbf{W}_1)  \\
 & = \prod\limits_{k}^{} \mathrm{Q}_{M}(w_k \mid \textbf{T}_t, \text{ }\Phi(\left( \frac{1}{K}\sum\limits_{m=1}^{K} {v}_m \right),  \left( \sum\limits_{m=1}^{N} v_m \alpha_{m}\right), \text{ }S ) )  \\
 & \text{ }\text{ }\text{ } \prod\limits_{t}^{} \mathrm{Q}_{F}(\textbf{T}_t \mid f_1({v}_1. \ldots, {v}_K, S), \text{ } f_2({v}_1. \ldots, {v}_K, S))  
\end{split} 
\end{equation}
using the weights of the LSTM in the architecture is denoted as $\textbf{W}_{L_1}$, $w_i$ as words of sentences, $v_i$ as regional image features, ${a}_iv_m$ as intermediate learnt parameters, $\mathrm{Q}_M(.)$ and $\mathrm{Q}_F(.)$ are the Image Caption generator and feature generator function respectively. $\mathrm{Q}_F(.)$ derives $\textbf{T}_i$ from $\textbf{v}$ of $\textbf{I}$, $S$ is the semantic feature map.

\begin{figure*}
\centering 
\includegraphics[width=\textwidth]{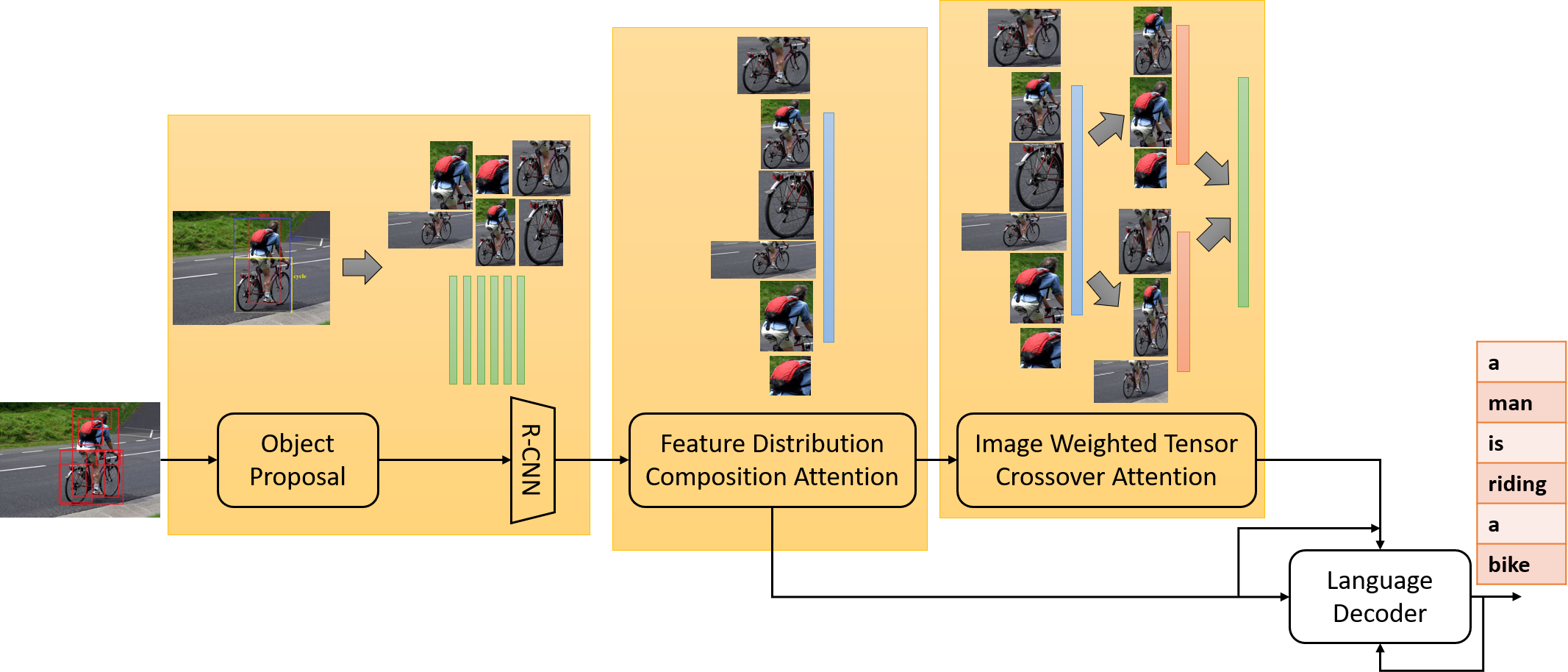}  
\caption{Overall Semi-FDC-Factorized-FDC-MRRC Architecture With FDC, IWTC and RCNN Feature Fusion.}
\label{fig:fdcV1}
\end{figure*}

\subsection{Semi-FDC-Factorized-FDC-MRRC}
Semi Feature Distribution Composition - Multiple Role Representation Crossover (Semi-FDC-Factorized-FDC-MRRC) provided another architecture, where we explored Dynamic Factorization techniques along with MRRC with the regional image features. However, compared to Semi-FDC-Factorized-Semantic-MRRC, this model was outperformed due to the absence of semantic distribution component $S$. 
Figure \ref{fig:fdcV1} provided the diagram for the Semi-FDC-Factorized-FDC-MRRC model, used for the experiments. The mathematical representation of Semi-FDC-Factorized-FDC-MRRC is presented as the followings.
\begin{equation}
 \overline{v} = \frac{1}{k} \sum\limits_{i=1}^{i=k} v_i
\end{equation} 
\begin{equation}
 \textbf{h}_{0}, \textbf{ } \textbf{c}_{0} = \textbf{W}_{h_0}\overline{v}, \textbf{W}_{c_0}\overline{v}
\end{equation}
\begin{equation} \label{eq:st5}
 \textbf{a}_{t} = \textbf{W}_{a} \tanh (\textbf{W}_{h} \textbf{h}_{t-1})
\end{equation}
\begin{equation}
 \alpha_t = \mathrm{softmax}(\textbf{a}_t)
\end{equation}
\begin{equation} \label{eq:en5}
 \textbf{q}_{t} = \widehat{v}_t = \sum\limits_{i=1}^{i=k} v_i \alpha_{i,t}
\end{equation}
\begin{equation}
 \textbf{p}_{t} = \textbf{W}_e \textbf{x}_{t-1}
\end{equation}
There are $\textbf{q}_{t}$ components for the MRRC representation $\textbf{T}_{t}$ equation, while the rest Language Decoder equations are same as Semi-FDC-Factorized-Semantic-MRRC.
\begin{equation}
\begin{split}
 \textbf{T}_{t} & =  
 \textbf{W}_{s_{12}}\text{ }\sigma(\textbf{W}_{s_{11}}f(\textbf{q}_{t}) + \textbf{W}_{w_{_{1}}}\sum\limits_{i=0}^{t-1} \textbf{W}_e \textbf{x}_{i} + \textbf{b}_1) 
 \text{ }\otimes \\
 & \tanh (\textbf{W}_{s_{22}} ( \textbf{q}_{t} \text{ }\sigma(\textbf{W}_{s_{21}}f(\textbf{q}_{t}) + \textbf{W}_{w_{_{2}}}\sum\limits_{i=0}^{t-1} \textbf{W}_e \textbf{x}_{i} + \textbf{b}_2) ) + \textbf{b}_3) \\
 & =  
 \textbf{W}_{s_{12}}\text{ }\sigma(\textbf{W}_{s_{11}}\textbf{h}_{t-1} + \textbf{W}_{w_{_{1}}}\sum\limits_{i=0}^{t-1} \textbf{W}_e \textbf{x}_{i} + \textbf{b}_1) 
 \text{ }\otimes \\
 & \tanh (\textbf{W}_{s_{22}} ( \textbf{q}_{t} \text{ }\sigma(\textbf{W}_{s_{21}}\textbf{h}_{t-1} + \textbf{W}_{w_{_{2}}}\sum\limits_{i=0}^{t-1} \textbf{W}_e \textbf{x}_{i} + \textbf{b}_2) ) + \textbf{b}_3) 
\end{split}
\end{equation}
\begin{equation} \label{eq:tag5}
\textbf{q}_{t,n} = \textbf{W}_{h,m} f(\textbf{q}_{t-1}) \odot \textbf{W}_{h,n} \textbf{q}_{t} = \textbf{W}_{h,m} \textbf{h}_{t-1} \odot \textbf{W}_{h,n} \textbf{q}_{t}
\end{equation} 
\begin{equation}
 \textbf{i}_{t} = \sigma(\textbf{W}_{pi}\textbf{p}_{t} + \textbf{W}_{qi}\textbf{q}_{t,n} + \textbf{W}_{Ti}\textbf{T}_{t} + \textbf{b}_{i})
\end{equation}
\begin{equation}
 \textbf{f}_{t} = \sigma(\textbf{W}_{pf}\textbf{p}_{t} + \textbf{W}_{qf}\textbf{q}_{t,n} + \textbf{W}_{Tf}\textbf{T}_{t} + \textbf{b}_{f})
\end{equation}
\begin{equation}
 \textbf{o}_{t} = \sigma(\textbf{W}_{po}\textbf{p}_{t} + \textbf{W}_{qo}\textbf{q}_{t,n} + \textbf{W}_{To}\textbf{T}_{t} + \textbf{b}_{o})
\end{equation}
\begin{equation}
 \textbf{g}_{t} = \tanh(\textbf{W}_{pg}\textbf{p}_{t} + \textbf{W}_{qg}\textbf{q}_{t,n} + \textbf{W}_{Tg}\textbf{T}_{t} + \textbf{b}_{g})
\end{equation}
\begin{equation}
 \textbf{c}_{t} = \textbf{f}_{t} \odot \textbf{c}_{t-1} + \textbf{i}_{t} \odot \textbf{g}_{t} 
\end{equation}
\begin{equation}
 \textbf{h}_{t} = \textbf{o}_{t} \odot \tanh(\textbf{c}_{t})
\end{equation}
where $\textbf{h}_{t}$ is pre-likelihood estimation of the topological sequence and is decoded as $\textbf{x}_t$ as in Equation \ref{eq:lang_decode}.

Mathematically, Semi Feature Distribution Composition - Multiple Role Representation Crossover (Semi-FDC-Factorized-FDC-MRRC), denoted as $f_{_{FDC_5}}(.)$, can be described as the followings probability distribution estimation.
\begin{equation} \label{eq:itb5}
\begin{split}
 f & _{_{FDC_4}}(\textbf{I}) = \prod\limits_{k}^{} \mathrm{Pr}(w_k \mid \textbf{T}_t, \text{ } \Phi({v}_1. \ldots, {v}_K), \text{ }S, \text{ } \textbf{W}_{L_1}) \\
 & \prod\limits_{t}^{} \mathrm{Pr}(\textbf{T}_t \mid f_1({v}_1. \ldots, {v}_K),  \text{ }f_2({v}_1. \ldots, {v}_K), \text{ }\textbf{W}_1)  \\
 & = \prod\limits_{k}^{} \mathrm{Pr}(w_k \mid \textbf{T}_t, \text{ }\Phi(\left( \frac{1}{K}\sum\limits_{m=1}^{K} {v}_m \right), \\
 &  \left( \textbf{W}_x \textbf{h}_{t-1} \sum\limits_{m=1}^{N} v_m \alpha_{m} \text{ } \textbf{W}_y \right) , \text{ }S ), \text{ } \textbf{W}_{L_1}  )  \\
 &      \text{ }\text{ }\text{ } \prod\limits_{t}^{} \mathrm{Pr}(\textbf{T}_t \mid f_1({v}_1. \ldots, {v}_K), \text{ } f_2({v}_1. \ldots, {v}_K),  \text{ }\textbf{W}_1)  \\
 & = \prod\limits_{k}^{} \mathrm{Q}_{M}(w_k \mid \textbf{T}_t, \text{ }\Phi(\left( \frac{1}{K}\sum\limits_{m=1}^{K} {v}_m \right),  \left( \sum\limits_{m=1}^{N} v_m \alpha_{m}\right), \text{ }S ) )  \\
 & \text{ }\text{ }\text{ } \prod\limits_{t}^{} \mathrm{Q}_{F}(\textbf{T}_t \mid f_1({v}_1. \ldots, {v}_K), \text{ } f_2({v}_1. \ldots, {v}_K))  
\end{split} 
\end{equation}
using the weights of the LSTM in the architecture is denoted as $\textbf{W}_{L_1}$, $w_i$ as words of sentences, $v_i$ as regional image features, ${a}_iv_m$ as intermediate learnt parameters, $\mathrm{Q}_M(.)$ and $\mathrm{Q}_F(.)$ are the Image Caption generator and feature generator function respectively. $\mathrm{Q}_F(.)$ derives $\textbf{T}_i$ from $\textbf{v}$ of $\textbf{I}$, $S$ is the semantic feature map.

\subsection{Reinforcement Learning}
There are hardly any improvement with reinforcement learning \cite{sur2019ucrlf} for bigger model as these architectures as the one-point error gradient can hardly provide scope of better optima for a system with so many variable weights. However, we did get some improvement for the best performing model Semi-FDC-Factorized-Semantic-MRRC. We define the Self-critical Sequence Training (SCST) \cite{rennie2017self} as Equation \ref{eq:CR1} or more precisely as Equation \ref{eq:IR}, where it utilizes the gradient of rewards associated with the sequences of generation for optimization and when the neural network is optimized with this reward difference, the procedure is known to be reinforcement learning, though the functional changes in neural network is not equivalent to the probability function determination. Mathematically, for image caption problem, the gradient of reward for reinforcement learning is defined as, 
\begin{equation} \label{eq:CR1}
 \frac{\delta L(\textbf{w})}{\delta \textbf{w}} = -\frac{1}{2b}\gamma \sum\limits_i \Phi(\textbf{y},\textbf{y}')
\end{equation}
\begin{equation} \label{eq:IR}
 \frac{\delta L(\textbf{w})}{\delta \textbf{w}} = -\frac{1}{2b}\gamma \sum\limits_i \Phi(\{y_1,\ldots,y_{c_1}\}, \{y'_1,\ldots,y'_{c_2}\})
\end{equation}
where $\Phi(.)$ is the evaluation function or the reward function that evaluates certain aspects of the generated captions $\{y_1,\ldots,y_c\} \in \textbf{y}'$ and the baseline captions $\{y'_1,\ldots,y'_c\} \in \textbf{y}$ and $c_1$ and $c_2$ is the length of the captions considered.

\section{Results \& Analysis} \label{section:results}
We performed different experiments with this data with varied dimension and different initialization approaches, along with different fine-tuning possibilities. The learning characteristics were better with $2-e$ learning rate while the best initialization strategy is with xavier initialization. $\text{Normal}(0, \text{ } 0.05)$ strategy of initialization worked in many of our experiments, but it is time consuming and may be required to run the program for more than 20 epochs for better results. With resource and time constraints, we find xavier initialization had the best strategy, where the initialization is considered with a distribution spanned over the whole layer instead of the network. We used 100 dimension word embedding but got this transformed from 300 dimension word embedding from GloVe. The image features and the RCNN features used were derived from ResNet architecture, and we used the already extracted features from \cite{Gan2016} and \cite{anderson2018bottom} works. 

\subsection{Dataset Description}
In this part, we will provide some insights of the MSCOCO data and different tricks for training. MSCOCO consists of 123287 train images and 566747 train sentence, where each image is associated with at least five sentences from a vocabulary of 8791 words. There are 5000 images (with 25010 sentences) for validation and 5000 images (with 25010 sentences) for testing. We used the same data split as described in Karpathy et al \cite{Karpathy2015Deep}. For SCN network, two sets of image features are being used: one is ResNet features with 2048 dimension feature vector and another is a MLP transformed representation called tag features with feature vector of 999 dimension and consisted of the probability of occurrence for the most appearing set of tags as incidents in the dataset. 

\subsection{Training \& Dimension Description}
We trained the models, once at a time, mainly due to limited resources and establishing the concepts are more important than trying to outperform a model, whose data is not publicly available. we tried various combination of dimension, but found that 1024 was best and our K80 GPU can handle it with feasible time for experiments. However, there is very narrow improvement between the 512 and 1024 dimension model. We ran the experiments for around 20/25 epochs and stopped when there was no significant improvement on validation set.

\subsection{Quantitative Analysis}
Numerical results shows that our model outperformed much better than any of the older works, considering that we are using their extracted data. Gradual improvement in feature extraction will provide better improvements. There are many works being done, where they report improvements but never share the data. Our model is both theoretically and practically robust and given any data, it has the potential to outperform the base model. Table \ref{table:PerformWithRL} provided the comparison of our results with all the previous works in this sector. We have only considered the works, where the data are available for comparison. Later, in Table \ref{table:PerformWithoutRL}, we have provided another similar comparison, but with  Reinforcement Learning technique for performance enhancement. However, Reinforcement Learning techniques like SCST never guarantee better performance, it keeps some scope open for experiments for these kinds of heuristic gradient to pull down the optimum to better optima, that can generate better captions based on a single comparison metrics. 

\begin{table*}
\centering
\caption{Performance Evaluation And Comparison Between Different Architectures without Reinforcement Learning}
\begin{tabular}{|c|c|c|c|c|c|c|c|c|}
\hline
 Algorithm & CIDEr-D & Bleu\_4 & Bleu\_3 & Bleu\_2 & Bleu\_1 & ROUGE\_L & METEOR  &  SPICE \\ 
\hline \hline 
 Human \cite{wu2017image} & 0.85 & 0.22 & 0.32 & 0.47 & 0.66 & 0.48 & 0.2 & -- \\ 
    Neural Talk \cite{Karpathy2015Deep} & 0.66 & 0.23 & 0.32 & 0.45 & 0.63 & -- & 0.20 & --  \\ 
    Mind’sEye \cite{Chen2015Mind} & -- & 0.19 & -- & -- & -- & -- & 0.20 & --  \\
    Google \cite{vinyals2015show}  & 0.94 & 0.31 & 0.41 & 0.54 & 0.71 & 0.53 & 0.25 & --  \\
    LRCN \cite{Donahue2015Long-term} & 0.87 & 0.28 & 0.38 & 0.53 & 0.70 & 0.52 & 0.24 & --  \\   
    Montreal \cite{Xu2015Show} & 0.87 & 0.28 & 0.38 & 0.53 & 0.71 & 0.52 &  0.24 & -- \\ 
    m-RNN \cite{Mao2014deep}  & 0.79 & 0.27 & 0.37 & 0.51 & 0.68 & 0.50 & 0.23 & --  \\
    \cite{Jia2015}  & 0.81 & 0.26 & 0.36 & 0.49 & 0.67 & -- & 0.23 & --  \\
    MSR \cite{Fang2015captions} & 0.91 & 0.29 & 0.39 & 0.53 & 0.70 & 0.52 & 0.25 & --  \\
    \cite{Jin2015Aligning} & 0.84 & 0.28 & 0.38 & 0.52 & 0.70 & -- & 0.24 & --  \\
    bi-LSTM \cite{wang2018image} & -- & 0.244 & 0.352 & 0.492 & 0.672 & -- & -- & --  \\ 
    MSR Captivator \cite{Devlin2015Language} & 0.93 & 0.31 & 0.41 & 0.54 & 0.72 & 0.53 & 0.25 & -- \\
    Nearest Neighbor \cite{devlin2015exploring} & 0.89 & 0.28 & 0.38 & 0.52 & 0.70 & 0.51 & 0.24 & -- \\
    MLBL \cite{kiros2014multimodal} & 0.74 & 0.26 & 0.36 & 0.50 & 0.67 & 0.50 & 0.22 & -- \\
    ATT \cite{You2016Image} & 0.94 & 0.32 & 0.42 & 0.57 & 0.73 & 0.54 & 0.25 & -- \\
    \cite{wu2017image} & 0.92 & 0.31 & 0.41 & 0.56 & 0.73 & 0.53 & 0.25 & -- \\
    \hline\hline
    Adaptive \cite{lu2017knowing}  & 1.085 & 0.332 & 0.439 & 0.580 & 0.742 & -- & 0.266 & --  \\
    MSM \cite{yao2017boosting}  & 0.986 & 0.325 & 0.429 & 0.565 & 0.730 & -- & 0.251 & --  \\ 
    ERD \cite{yang1605encode}  & 0.895 & 0.298 & -- & -- & -- & -- & 0.240 & --  \\ 
    Att2in \cite{rennie2017self}  & 1.01 & 0.313 & -- & -- & -- & -- & 0.260 & --  \\     
    NBT \cite{lu2018neural}  & 1.07 & 0.347 & -- & -- & 0.755 & -- & 0.271 & 0.201  \\ 
    Attribute-Attention \cite{chen2018show} & 1.044 & 0.338 & 0.443 & 0.579 & 0.743 & 0.549  & -- & -- \\ %
    LSTM \cite{Gan2016} & 0.889 & 0.292 & 0.390 & 0.525 & 0.698 & -- & 0.238 & -- \\
    SCN \cite{Gan2016} & 1.012 & 0.330 & 0.433 & 0.566 & 0.728 & -- & 0.257 & --  \\
    Up-Down$^{**}$ \cite{anderson2018bottom}  & 1.054 & 0.334 & -- & -- & 0.745 & 0.544 & 0.261 & 0.192  \\ %
    Up-Down$\dagger$ \cite{anderson2018bottom}  & 1.135 & 0.362 & -- & -- & 0.772 & 0.564 & 0.270 & 0.203  \\ %
 \hline \hline

 FDC-MRRC & 1.073 & 0.351 & 0.459 & 0.595 & 0.752 & 0.557 & 0.265 & 0.198 \\ 
 
 Semi-Factorized-FDC-MRRC & 1.075 & 0.349 & 0.456 & 0.591 & 0.748 & 0.557 & 0.265 & 0.195 \\ 
 
 Full-Factorized-FDC-MRRC & 1.078 & 0.349 & 0.456 & 0.592 & 0.751 & 0.556 & 0.266 & 0.196 \\ 
 
 Semi-FDC-Factorized-Semantic-MRRC & \textbf{1.080} & \textbf{0.352} & \textbf{0.460} & \textbf{0.598} & \textbf{0.755} & \textbf{0.559} & \textbf{0.265} & \textbf{0.197} \\ 
 
 Semi-FDC-Factorized-FDC-MRRC & 1.073 & 0.348 & 0.453 & 0.589 & 0.749 & 0.554 & 0.265 & 0.195 \\ \hline
 
 \multicolumn{9}{l}{\textsuperscript{**}\footnotesize{Visual Feature from ResNet Architecture is used}} \\ 
 \multicolumn{9}{l}{\textsuperscript{$\dagger$}\footnotesize{R-CNN Visual Feature from ResNet Architecture is Used}}
\end{tabular}
\label{table:PerformWithoutRL}
\end{table*}

\begin{table*}
\centering
\caption{Performance Evaluation And Comparison Between Different Architectures with Reinforcement Learning Fine-tuning}
\begin{tabular}{|c|c|c|c|c|c|c|c|c|}
\hline
 Algorithm & CIDEr-D & Bleu\_4 & Bleu\_3 & Bleu\_2 & Bleu\_1 & ROUGE\_L & METEOR  &  SPICE \\ 
\hline \hline
    Adaptive \cite{lu2017knowing}  & 1.085 & 0.332 & 0.439 & 0.580 & 0.742 & -- & 0.266 & --  \\
    MSM \cite{yao2017boosting}  & 0.986 & 0.325 & 0.429 & 0.565 & 0.730 & -- & 0.251 & --  \\ 
    ERD \cite{yang1605encode}  & 0.895 & 0.298 & -- & -- & -- & -- & 0.240 & --  \\ 
    Att2in \cite{rennie2017self}  & 1.01 & 0.313 & -- & -- & -- & -- & 0.260 & --  \\  
    NBT \cite{lu2018neural}  & 1.07 & 0.347 & -- & -- & 0.755 & -- & 0.271 & 0.201  \\ 
    \cite{chen2018show} & 1.044 & 0.338 & 0.443 & 0.579 & 0.743 & 0.549  & -- & -- \\ %
    LSTM \cite{Gan2016} & 0.889 & 0.292 & 0.390 & 0.525 & 0.698 & -- & 0.238 & -- \\
    SCN \cite{Gan2016} & 1.012 & 0.330 & 0.433 & 0.566 & 0.728 & -- & 0.257 & --  \\
    Up-Down$^{**}$ \cite{anderson2018bottom}  & 1.054 & 0.334 & -- & -- & 0.745 & 0.544 & 0.261 & 0.192  \\ %
    Up-Down$\dagger$ \cite{anderson2018bottom}  & 1.135 & 0.362 & -- & -- & 0.772 & 0.564 & 0.270 & 0.203  \\ %
    Up-Down$^{**}$ + RL \cite{anderson2018bottom}   & 1.111 & 0.340 & -- & -- & 0.766 & 0.549 & 0.265 & 0.202   \\ %
    Up-Down$\dagger$ + RL \cite{anderson2018bottom} & 1.201 & 0.363 & -- & -- & 0.798 & 0.569 & 0.277 & 0.214 \\ %
 \hline \hline
 FDC-MRRC & 1.078 & 0.351 & 0.457 & 0.598 & 0.754 & 0.557 & 0.267 & 0.199 \\ 
 
 Semi-Factorized-FDC-MRRC & 1.079 & 0.350 & 0.457 & 0.593 & 0.749 & 0.556 & 0.265 & 0.195 \\ 
 
 Full-Factorized-FDC-MRRC & 1.075 & 0.351 & 0.457 & 0.593 & 0.751 & 0.555 & 0.266 & 0.196 \\ 
 
 Semi-FDC-Factorized-Semantic-MRRC & \textbf{1.082} & \textbf{0.353} & \textbf{0.460} & \textbf{0.597} & \textbf{0.753} & \textbf{0.557} & \textbf{0.266} & \textbf{0.197} \\ 
 
 Semi-FDC-Factorized-FDC-MRRC & 1.075 & 0.351 & 0.457 & 0.592 & 0.749 & 0.557 & 0.264 & 0.196 \\ \hline
 \multicolumn{9}{l}{\textsuperscript{**}\footnotesize{Visual Feature from ResNet Architecture is used}} \\ 
 \multicolumn{9}{l}{\textsuperscript{$\dagger$}\footnotesize{R-CNN Visual Feature from ResNet Architecture is Used}}
\end{tabular}
\label{table:PerformWithRL}
\end{table*}

\subsection{Qualitative Analysis}
\begin{figure*}[t] 
\centering 
\includegraphics[width=\textwidth]{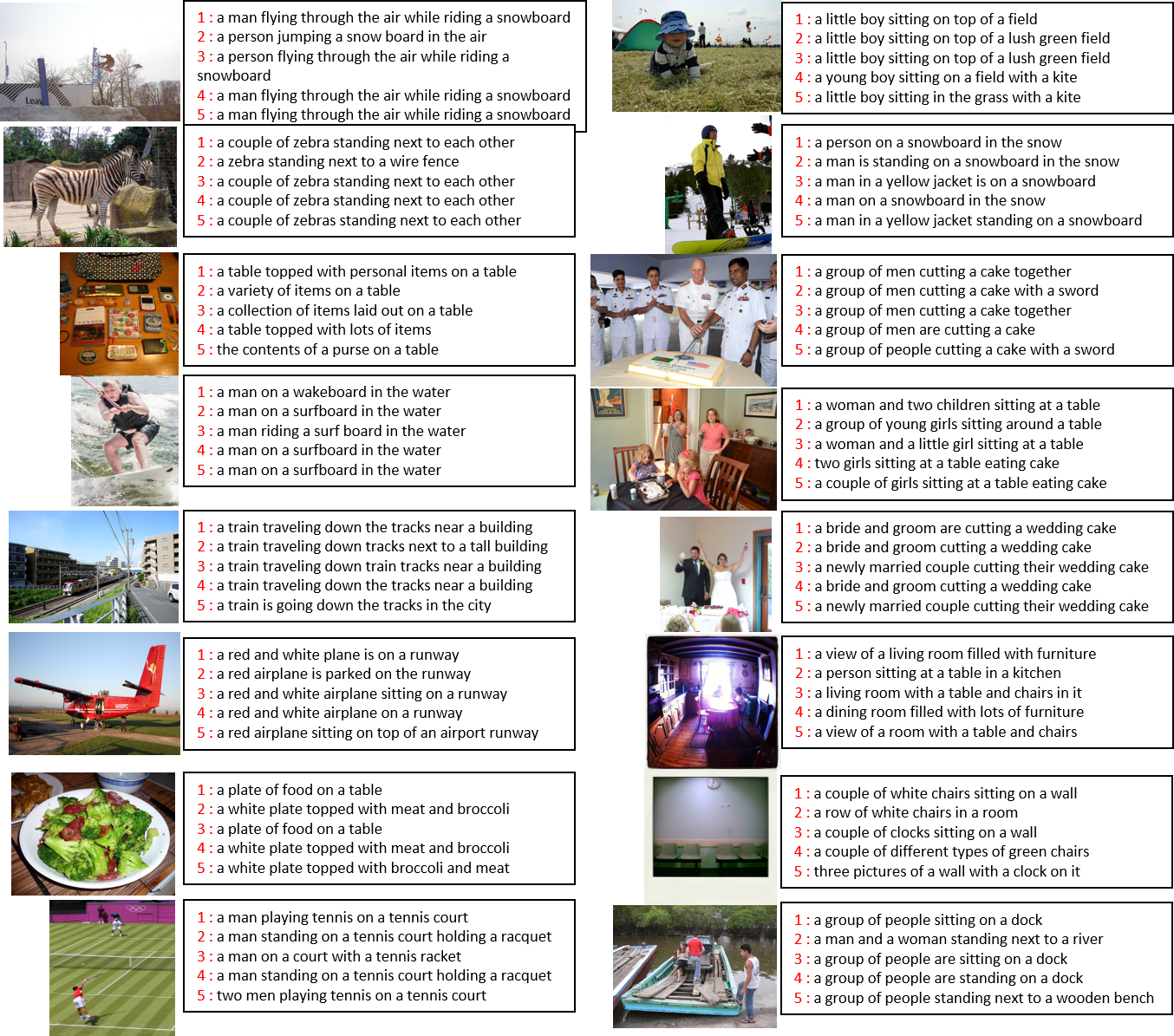}  
\caption{Qualitative Analysis. Part 1. Here, 1 $\rightarrow$ Semi-FDC-Factorized-Semantic-MRRC :  2 $\rightarrow$ Semi-FDC-Factorized-FDC-MRRC  3 $\rightarrow$ FDC-MRRC 4 $\rightarrow$ Semi-Factorized-FDC-MRRC 5 $\rightarrow$ Full-Factorized-FDC-MRRC }
\label{fig:QualitativeAnalysis1}
\end{figure*}

\begin{figure*}[!h] 
\centering 
\includegraphics[width=\textwidth]{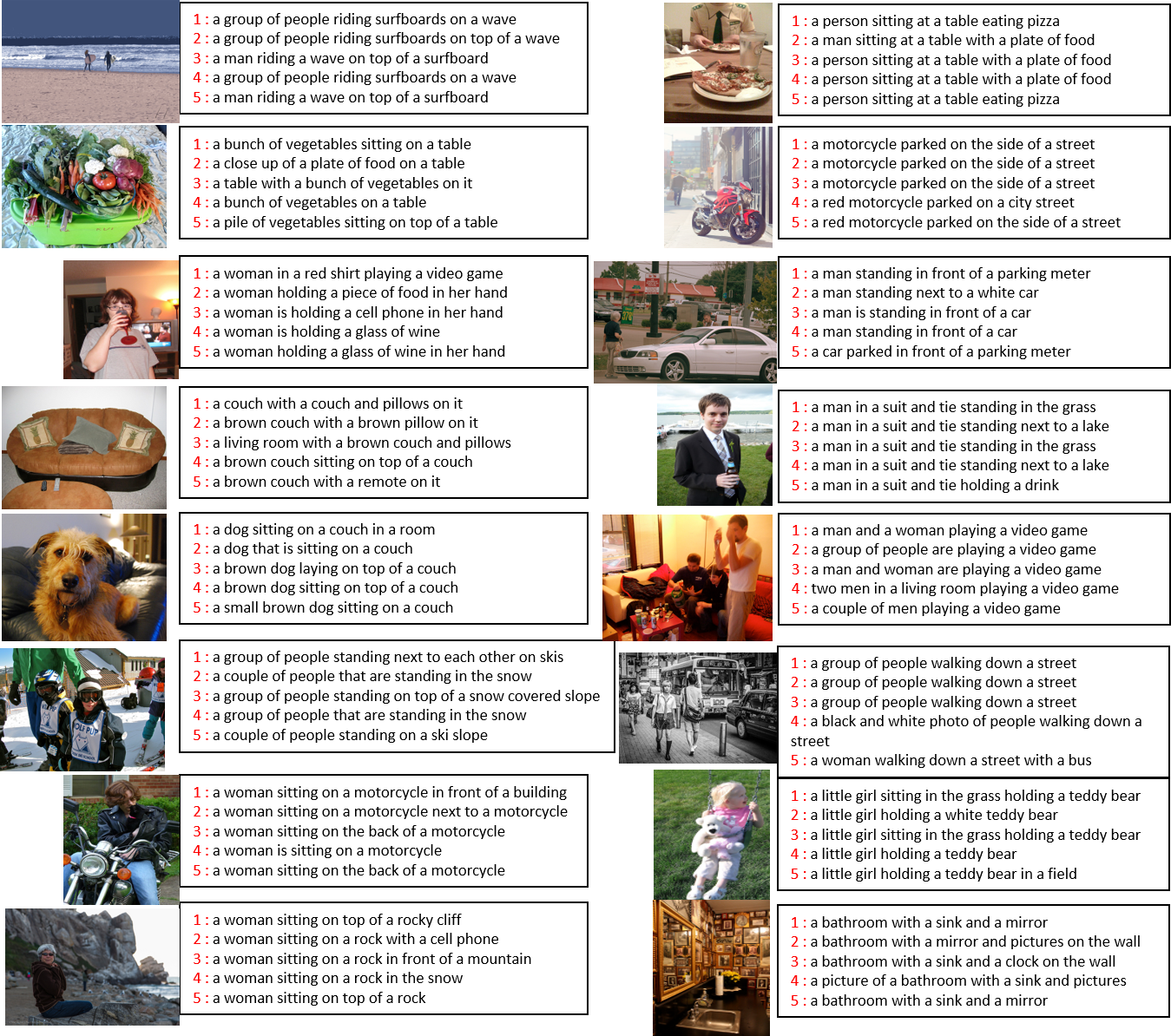}  
\caption{Qualitative Analysis. Part 2. Here, 1 $\rightarrow$ Semi-FDC-Factorized-Semantic-MRRC :  2 $\rightarrow$ Semi-FDC-Factorized-FDC-MRRC  3 $\rightarrow$ FDC-MRRC 4 $\rightarrow$ Semi-Factorized-FDC-MRRC 5 $\rightarrow$ Full-Factorized-FDC-MRRC }
\label{fig:QualitativeAnalysis2}
\end{figure*}

Quantitative analysis can never the perfect judge for language generator, though the spectrum of the context can be judged statistically. Hence, we have included some instances of the generated captions, which has shown significant improvement in quality of the descriptions. Figure \ref{fig:QualitativeAnalysis1} and Figure \ref{fig:QualitativeAnalysis2} provided some instances of generated captions using all the architectures. 
It is difficult to judge from numerical results (in Table \ref{table:PerformWithoutRL} or Table \ref{table:PerformWithRL}) if a model is better than the other and didn't improved captions. The following qualitative analysis will reflect some of them as it is very difficult to conclude such claims from the quantitative numerical figures related to languages. 
Most of the caption generated work used diverse evaluation methods for the generated captions, but there are requirement of analysis of quality and grammatically correctness.
There are captions which are true is some sense, but due to the limited proficiency of the language evaluation models, it has unable to capture them. 
Figure \ref{fig:QualitativeAnalysis1} and Figure \ref{fig:QualitativeAnalysis2} provided qualitative instances among various models.  

\section{Discussion} \label{section:discussion}
In this work, we analyzed the prospect of fusion of tensor product based structuring with semantic tag feature and RCNN features and demonstrated that these three combination can work well and create a state-of-the-art architecture for caption generation through adopting better representation generation and the ability to segregate the attrubutional features of the objects and activities from the images to the sentences. This architecture has adopted several existing features being offered through transfer learning and utilized these into a coherent model that is sensitive to different variations of the representation and thus help in better and longer sentence generation. 
The main contribution of this work is marked by the analysis of interaction of semantic features, RCNN features and tensor product. While RCNN is focused on regional objects, semantic features are focused on creation of a combination of objects and interaction useful for decoding of the word sequence, and tensor product is used for orthogonal combination of the features that can be decoded for the segregation of the visual features. We were being able to create high performance with these combination and out-performed some of the present works.

\ifCLASSOPTIONcaptionsoff
  \newpage
\fi

%








\end{document}